\documentclass[10pt,twocolumn,letterpaper]{article}

\usepackage[pagenumbers]{cvpr}      

\usepackage[accsupp]{axessibility} 
\usepackage{graphicx}
\usepackage{amsmath}
\usepackage{amssymb}
\usepackage{booktabs}

\usepackage{tablefootnote}
\usepackage{subcaption}
\usepackage{graphicx}
\usepackage{verbatim}
\usepackage{color}
\usepackage{booktabs}
\usepackage{float}
\usepackage{enumitem}

\usepackage[table]{xcolor}

\usepackage{caption}
\newcommand{\figref}[1]{Fig.~\ref{#1}}
\newcommand{\secref}[1]{Sec.~\ref{#1}}
\newcommand{\tabref}[1]{Table~\ref{#1}}

\usepackage[pagebackref,breaklinks,colorlinks]{hyperref}

\usepackage[capitalize]{cleveref}
\crefname{section}{Sec.}{Secs.}
\Crefname{section}{Section}{Sections}
\Crefname{table}{Table}{Tables}
\crefname{table}{Tab.}{Tabs.}

\def\cvprPaperID{4667} 
\def\confName{CVPR}
\def\confYear{2022}

\begin{document}

\title{Learning Local Displacements for Point Cloud Completion}

\author{Yida Wang\textsuperscript{1}\\
{\tt\small yida.wang@tum.de}\\
\textsuperscript{1}Technische Universit\"at M\"unchen
\and
David Joseph Tan\textsuperscript{2}\\
{\tt\small djtan@google.com}\\

\and
Nassir Navab\textsuperscript{1}\\
{\tt\small nassir.navab@tum.de}\\

\and
Federico Tombari\textsuperscript{1,2}\\
{\tt\small tombari@in.tum.de}\\
}

\author{
Yida Wang\textsuperscript{1}, 
David Joseph Tan\textsuperscript{2}, 
Nassir Navab\textsuperscript{1}, 
Federico Tombari\textsuperscript{1,2}\\
\textsuperscript{1}Technische Universit\"at M\"unchen ~~~~~
\textsuperscript{2}Google Inc.
}

\maketitle

\begin{abstract}
We propose a novel approach aimed at object and semantic scene completion from a partial scan represented as a 3D point cloud. 
Our architecture relies on three novel layers that are used successively within an encoder-decoder structure and specifically developed for the task at hand. 
The first one carries out feature extraction by matching the point features to a set of pre-trained local descriptors.
Then, to avoid losing individual descriptors as part of standard operations such as max-pooling, we propose an alternative neighbor-pooling operation that relies on adopting the feature vectors with the highest activations. 
Finally, up-sampling in the decoder modifies our feature extraction in order to increase the output dimension.
While this model is already able to achieve competitive results with the state of the art, we further propose a way to increase the versatility of our approach to process point clouds. To this aim, we introduce a second model that assembles our layers within a transformer architecture.
We evaluate both architectures on object and indoor scene completion tasks, achieving state-of-the-art performance.

\end{abstract}

\section{Introduction}

\begin{figure}[!t]
\centering
\includegraphics[width=1.0\linewidth]{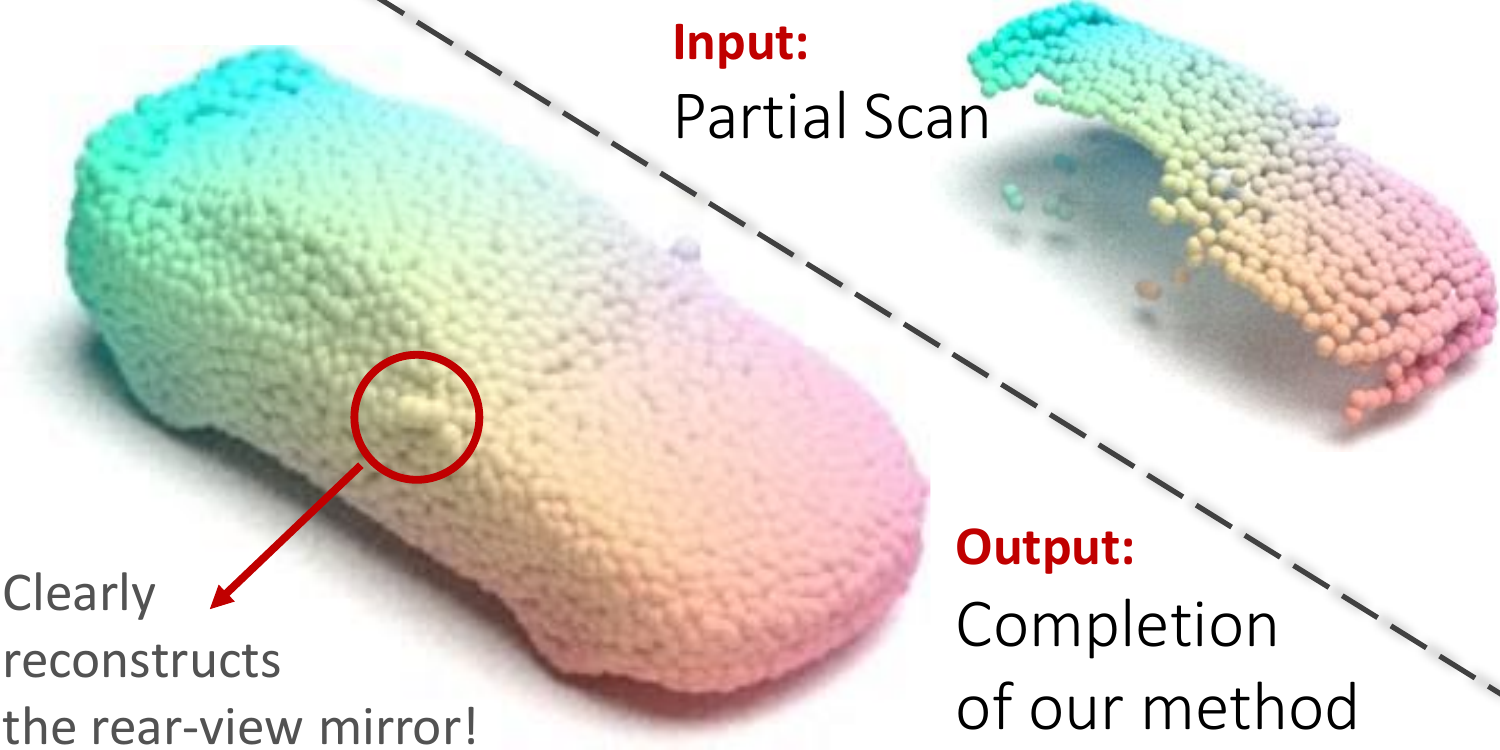}
\caption{From the input partial scan to our object completion, we visualize the amount of detail in our reconstruction.}
\label{fig:teaser}
\end{figure}

Understanding the entire 3D space is essential for both humans and machines to understand how to safely navigate an environment or how to interact with the objects around them.
However, when we capture the 3D structure of an object or scene from a certain viewpoint, a large portion of the whole geometry is typically missing due to self-occlusion and/or occlusion from its surrounding. 
To solve this problem, geometric completion of scenes~\cite{wang2019forknet, song2017semantic, cai2021semantic} and objects~\cite{yuan2018pcn, liu2020morphing, grnet_xie, yu2021pointr, pan2021variational} has emerged as a task that takes on a 2.5D/3D observation and fills out the occluded regions, as illustrated in \figref{fig:teaser}.

There are multiple ways to represent 3D shapes. 
Point cloud~\cite{chang2015shapenet, dai2017scannet}, volumetric grid~\cite{song2017semantic, dai2018scancomplete}, mesh~\cite{Groueix_2018_CVPR} and implicit surfaces~\cite{park2019deepsdf, occupancy_Mescheder, DISN19} are among the most common data formats. 
These representations are used for most 3D-related computer vision tasks such as segmentation, classification and completion.
For what concerns geometric completion, most works are focused on either point cloud or volumetric data. Among them, the characteristic of having an explicitly defined local neighbourhood makes volumetric data easier to process with 3D convolutions~\cite{dai2017shape, yang20173d, yang2018dense}. 
One drawback introduced by the predefined local neighborhood is the inaccuracy due to the constant resolution of the voxels, meaning that one voxel can represent several small structures.

On the other hand, 
point clouds have the advantage of not limiting the local resolution, although they come with their own sets of drawbacks. 
Mainly, there are two problems in processing point clouds: the undefined local neighborhood and unorganized feature map. 
Aiming at solving these issues, 
PointNet++~\cite{qi2017pointnet++}, PMP-Net~\cite{wen2020pmp}, PointConv~\cite{wu2019pointconv} and PointCNN~\cite{li2018pointcnn}
employ $k$-nearest neighbor search to define a local neighborhood, while PointNet~\cite{qi2017pointnet} and SoftPoolNet~\cite{wang_softpool} adopt the pooling operation to achieve permutation invariant features.
Notably, point cloud segmentation and classification were further improved by involving $k$-nearest neighbor search to form local features in PointNet++\cite{qi2017pointnet++} compared to global features in PointNet~\cite{qi2017pointnet}. 
Several variations of PointNet~\cite{qi2017pointnet} also succeeded in improving point cloud completion as demonstrated in FoldingNet~\cite{yang2018foldingnet}, PCN~\cite{yuan2018pcn}, MSN~\cite{liu2020morphing}.
Other methods such as SoftPoolNet~\cite{wang_softpool} and GRNet~\cite{grnet_xie} explicitly present local neighbourhood in sorted feature map and voxel space, respectively.

This paper investigates grouping local features to improve the point cloud completion of objects and scenes.
We apply these operation in encoder-decoder architectures which iteratively uses a feature extraction operation with the help of a set of displacement vectors as part of our parametric model. 
In addition, we also introduce a new pooling mechanism called neighbor-pooling, aimed at down-sampling the data in the encoder while, at the same time, preserving individual feature descriptors.
Finally, we propose a new loss function that gradually reconstructs the target from the observable to the occluded regions.
The proposed approach is evaluated on both object completion dataset with ShapeNet~\cite{chang2015shapenet}, and semantic scene completion on NYU~\cite{silberman2012indoor} and CompleteScanNet~\cite{wu2020scfusion}, attaining significant improvements producing high resolutions reconstruction with fine-grained details. 

\section{Related works}

This section focuses on the three most related fields -- point cloud completion, point cloud features and semantic scene completion.

\paragraph{Point cloud completion.}

Given the partial scan of an object similar to \figref{fig:teaser}, 3D completion aims at estimating the missing shape. In most cases, the missing region is due to self-occlusion since the partial scan is captured from a single view of the object.
Particularly for point cloud, 
FoldingNet~\cite{yang2018foldingnet} and 
AtlasNet~\cite{Groueix_2018_CVPR} 
are among the first works to propose an object completion based on PointNet~\cite{qi2017pointnet} features by deforming one or more 2D grids into the desired shape.
Then, PCN~\cite{yuan2018pcn} extended their work by deforming a collection of much smaller 2D grids in order to reconstruct finer structures. 

Through encoder-decoder architectures, ASFM-Net~\cite{xia2021asfm} and VRCNet~\cite{pan2021variational} match the encoded latent feature with a completion shape prior, which produce good coarse completion results. 
To preserve the observed geometry from the partial scan for the fine reconstruction, MSN~\cite{liu2020morphing} and VRCNet~\cite{pan2021variational} bypass the observed geometries by using either the minimum density sampling (MDS) or the farthest point sampling (FPS) from the observed surface and building skip connections. 
By embedding a volumetric sub-architecture, GRNet~\cite{grnet_xie} preserves the discretized input geometries with the volumetric U-connection without sampling in the point cloud space.
In more recent works, PMP-Net~\cite{wen2020pmp} gradually reconstructs the entire object from the observed to the nearest occluded regions. 
Also focusing on only predicting the occluded geometries, PoinTr~\cite{yu2021pointr} is among the first few transformer methods targeted on point cloud completion by translating the partial scan proxies into a set of occluded proxies to further refine the reconstruction.

\paragraph{Point cloud features.}

Notably, a large amount of work in object completion~\cite{yang2018foldingnet, Groueix_2018_CVPR, yuan2018pcn, liu2020morphing, grnet_xie, wang_softpool, wen2020pmp} rely on PointNet features~\cite{qi2017pointnet}. 
The main advantage of \cite{qi2017pointnet} is its capacity to be permutation invariant through max-pooling. 
This is a crucial characteristic for the input point cloud because its data is unstructured.

However, the max-pooling operation disassembles the point-wise features and ignores the local neighborhood in 3D space. 
This motivated SoftPoolNet~\cite{wang_softpool} to solve this problem by sorting the feature vectors based on the activation instead of taking the maximum values for each element.
In effect, they were able to concatenate the features to form a 2D matrix so that a traditional 2D convolution from CNN can be applied.

Apart from building feature representation through pooling operations, 
PointNet++~\cite{qi2017pointnet++} samples the local subset of points with the farthest point sampling (FPS) then feeds it into PointNet~\cite{qi2017pointnet}.
Based on this feature, SA-Net~\cite{Wen_2020_CVPR} then groups the features in different resolutions with KNN for further processing, while PMP-Net~\cite{wen2020pmp} uses PointNet++ features to identify the direction to which the object should be reconstructed.  PoinTr~\cite{yu2021pointr} also solves the permutational invariant problem without pooling by adding the positional coding of the input points into a transformer.

\paragraph{Semantic scene completion.}

All the point cloud completion are designed to reconstruct a single object.  
Extending these methods from objects to scenes is difficult because of the difference in size and content. 
When we tried to train these methods for objects, we noticed that the level of noise is significantly increased such that most objects in the scene are unrecognizable. 
Evidently, for semantic scene completion, the objective is not only to build the full reconstruction of the scene but also to semantically label each component. 

On the other hand, there have been a number of methods for semantic scene completion based on voxel grids that was initiated by SSCNet~\cite{song2017semantic}. 
Using a similar volumetric data with 3D convolutions~\cite{dai2017shape, yang20173d, yang2018dense}, VVNet~\cite{guo2018view} convolves on the 3D volumes which are back-projected from the depth images, revealing the camera view instead of a TSDF volume.
Later works such as 3D-RecGAN~\cite{yang20173d} and ForkNet~\cite{wang2019forknet} use discriminators to optimize the convolutional encoder and decoder during training.
Since 3D convolutions are heavy in terms of memory consumption especially when the input is presented in high resolution, SketchSSC~\cite{chen20203d} learns the 3D boundary of all objects in the scene to quickly estimate the resolution of the invariant features. 

Although there are quite many methods targeting on volumetric semantic scene completion, there are still no related works proposed explicitly for point cloud semantic scene completion which we achieved  in this paper.

\section{Operators}
\label{sec:operation}

\newcommand{\cloud}{\mathcal{P}}
\newcommand{\Pin}{\cloud_\text{in}}
\newcommand{\Pout}{\cloud_\text{out}}
\newcommand{\Pgt}{\cloud_\text{gt}}

Whether reconstructing objects or scenes from a single depth image, the objective is to process the given point cloud of the partial scan $\Pin$ to reconstruct the complete structure $\Pout$. 
Most deep learning solutions~\cite{yang2018foldingnet, yuan2018pcn, wang_softpool, pan2021variational, liu2020morphing} solve this problem by building an encoder-decoder architecture. The encoder takes the input point cloud to iteratively \emph{down}-sample it into its latent feature. Then, the decoder iteratively \emph{up}-sample the latent feature to reconstruct the object or scene.
In this section, we illustrate our novel down-sampling and up-sampling operations that cater to point cloud completion. Thereafter, in the following sections, we use our operators as building blocks to assemble two different encoder-decoder architectures that perform object completion and semantic scene completion. We also discuss the associated loss functions.

\subsection{Down-sampling operation}
\label{sec:down}

\newcommand{\feat}{\mathbf{f}}
\newcommand{\featset}{\mathcal{F}}
\newcommand{\real}{\mathbb{R}}
\newcommand{\Fin}{\featset_\text{in}}
\newcommand{\Fout}{\featset_\text{out}}
\newcommand{\Fup}{\featset_\text{up}}
\newcommand{\Fdown}{\featset_\text{down}}

\begin{figure}[!t]
\centering
\includegraphics[width=0.85\linewidth]{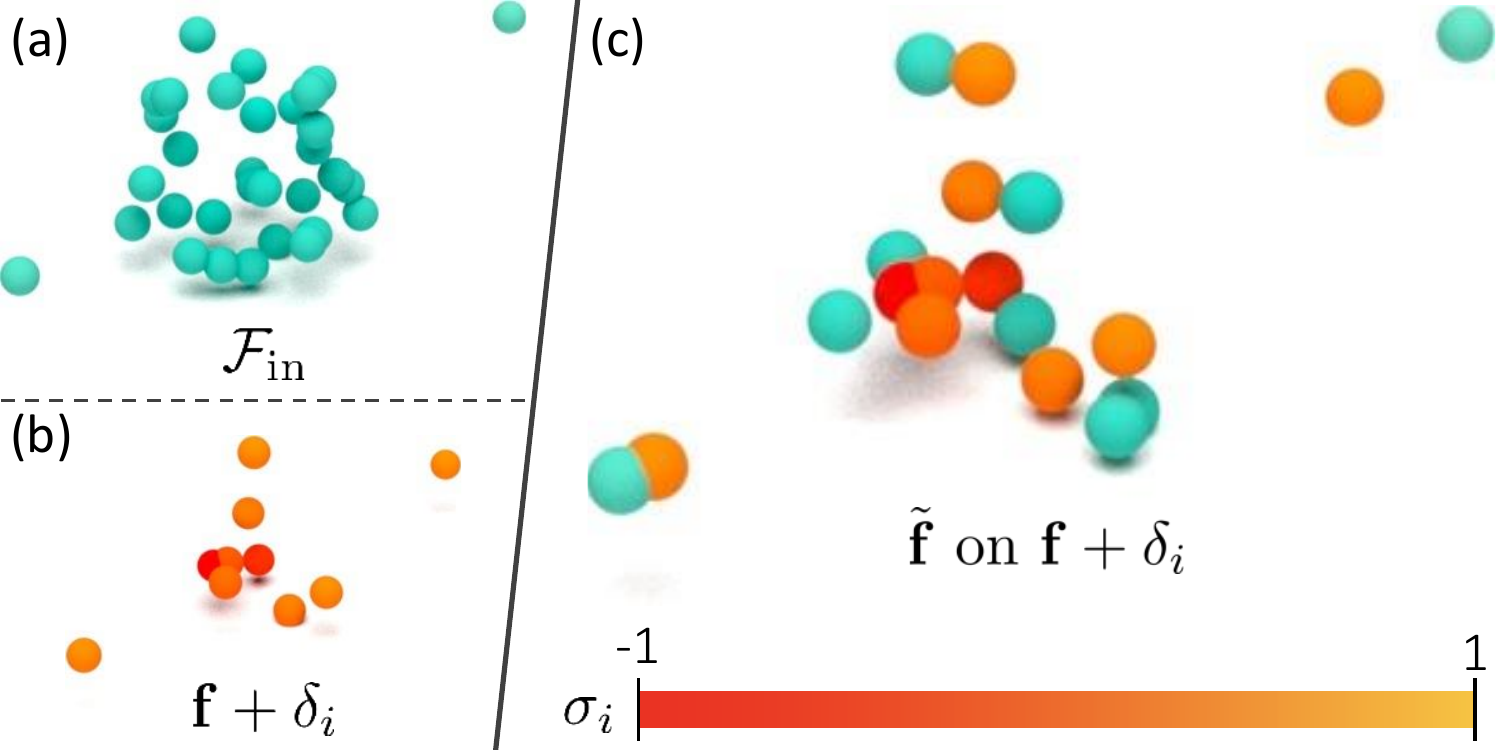}
\caption{(a) $k$-nearest neighbor in reference to an anchor $\feat$; (b) displacement vectors around the anchor $\feat + \delta_i$ and the corresponding weight $\sigma_i$; and, (c) closest features $\tilde{\feat}$ to $\feat + \delta_i$ for all $i$.
}
\label{fig:graph_conv}
\end{figure}

To formalize the down-sampling operation, we denote the input as the set of feature vectors $\Fin = \{\feat_i\}_{i=1}^{|\Fin|}$ where $\feat_i$ is a feature vector and $|\cdot|$ is the number of elements in the set.
Note that, in the first layer of the encoder, $\Fin$ is then set to the coordinates of the input point cloud.
We introduce a novel down-sampling operation inspired from the Iterative Closest Point (ICP) algorithm~\cite{besl1992method,chen1992object}. 
Taking an arbitrary anchor $\feat$ from $\Fin$, we start by defining a vector $\delta \in \real^{D_\text{in}}$. From the trainable variable $\delta$, we find the feature closest to $\feat+\delta$ and compute the distance. This is formally formulated as a function
\begin{align}
d \left( \feat, \delta \right) = 
\min_{\forall \tilde{\feat} \in \Fin}
\| (\feat + \delta) - \tilde{\feat} \| 
\label{eq:closest_distance}
\end{align}
where $\delta$ represents a displacement vector from $\feat$. 
Multiple displacement vectors are used to describe the local geometry, each with a weight $\sigma \in \real$. 
We then assign the set as $\{(\delta_i, \sigma_i)\}_{i=1}^s$ and aggregate them with the weighted function
\begin{align}
g ( \feat ) = 
\sum_{i=0}^{s}
\sigma_i \tanh{\frac{\alpha}{d(\feat, \delta_i) + \beta}} 
\label{eq:g_func}
\end{align}
where the constants $\alpha$ and $\beta$ are added for numerical stability.
Here, the hyperbolic tangent in $g(\feat)$ produces values closer to 1 when the distance $d(\cdot)$ is small and closer to 0 when the distance is large.
In practice, we can speed-up \eqref{eq:closest_distance} with the $k$-nearest neighbor search for each anchor.
A simple example of this operation is depicted in \figref{fig:graph_conv}. This illustrates the operation in the first layer where we process the point cloud so that we can geometrically plot a feature in $\Fin$ with respect to $\{(\delta_i, \sigma_i)\}_{i=1}^s$.

Furthermore, to enforce the influence of the anchor in this operation, we also introduce the function
\begin{align}
h ( \feat ) = 
\rho \cdot \feat
\end{align}
that projects $\feat$ on $\rho \in \real^{D_\text{in}}$, which is a trainable parameter. Note that both functions $g(\cdot)$ and $h(\cdot)$ produce a scalar value.

Thus, if we aim at building a set of output feature vectors, each with a dimension of $D_\text{out}$, we construct the set as 
\begin{align}
\Fout = \left\{ \left[
g_b(\feat_a) + h(\feat_a)
\right]_{b=1}^{D_\text{out}}
\right\}_{a=1}^{|\Fin|}
\label{eq:down}
\end{align}
where different sets of trainable parameters 
$\{(\delta_i,\sigma_i)\}_{i=1}^{s}$
are assigned to each element, while different $\rho$ for each output vector. Moreover, the variables $s$ in \eqref{eq:g_func} and $D_\text{out}$ in \eqref{eq:down} are the hyper-parameters. We label this operation as the \emph{feature extraction}.

It is noteworthy to mention that the proposed down-sampling operation is different from 3D-GCN~\cite{lin2020convolution}, which only takes the cosine similarity. While still being scale-invariant, hence suitable for object classification and segmentation, they ignore the metric structure of the local 3D geometry; consequently, making completion difficult because the original scale of the local geometry is missing.

\vspace{-5pt}\paragraph{Neighbor pooling.}

The final step in our down-sampling operation is to reduce the size of $\Fout$ with pooling.
However, unlike Graph Max-Pooling (GMP)~\cite{lin2020convolution}, that takes the element-wise maximum value of the feature across all the vectors, we select the subset of feature vectors with the highest activations. 
Therefore, while GMP disassembles their features as part of their pooling operation, we preserve the feature descriptors from $\Fout$.
From the definition of $\Fout$ in \eqref{eq:down}, we base our activation for each vector $\feat_a$ 
\begin{align}
\mathcal{A}_a = 
\sum_{b=1}^{D_\text{out}}
\tanh |g_b(\feat_a)|
\label{eq:activation}
\end{align}
on the results of $g(\cdot)$ from \eqref{eq:g_func}.
Thereafter, we only take the $\frac{1}{\tau}$ of the number of feature vectors with the highest activations.

\subsection{Up-sampling operation}
\label{sec:up}

The up-sampling and pooling operations in the encoder reduce the point cloud to a latent vector. In this case, if we directly use the operation in \eqref{eq:down}, the first layer in the decoder ends up with one vector since $|\featset_\text{in}|$ is one.
Subsequently, all the other layers in the decoder result in a single vector.
To solve this issue, our up-sampling iteratively runs \eqref{eq:down} so that, denoting $\Fin$ as the input to the layer, we build the set of output feature vectors as
\begin{align}
\Fup
&= \left\{ \Fout^u
\right\}_{u=1}^{ N_\text{up}}
\nonumber \\
&= \left\{ \left[
g_b^u(\feat_a) + h_b^u(\feat_a)
\right]_{b=1}^{D_\text{out}}
\right\}_{a=1, u=1}^{a = |\Fin|, u = N_\text{up}} 
\label{eq:up}
\end{align}
which increases the number of vectors by $N_\text{up}$. 
As a result, $\featset_\text{up}$ is a set of $N_u \cdot |\Fin|$ feature vectors.
In addition to the list of hyper-parameters in \secref{sec:down}, our up-sampling operation also takes $N_\text{up}$ as a hyper-parameter.

\begin{figure} [!t]
\centering
\includegraphics[width=0.85\linewidth]{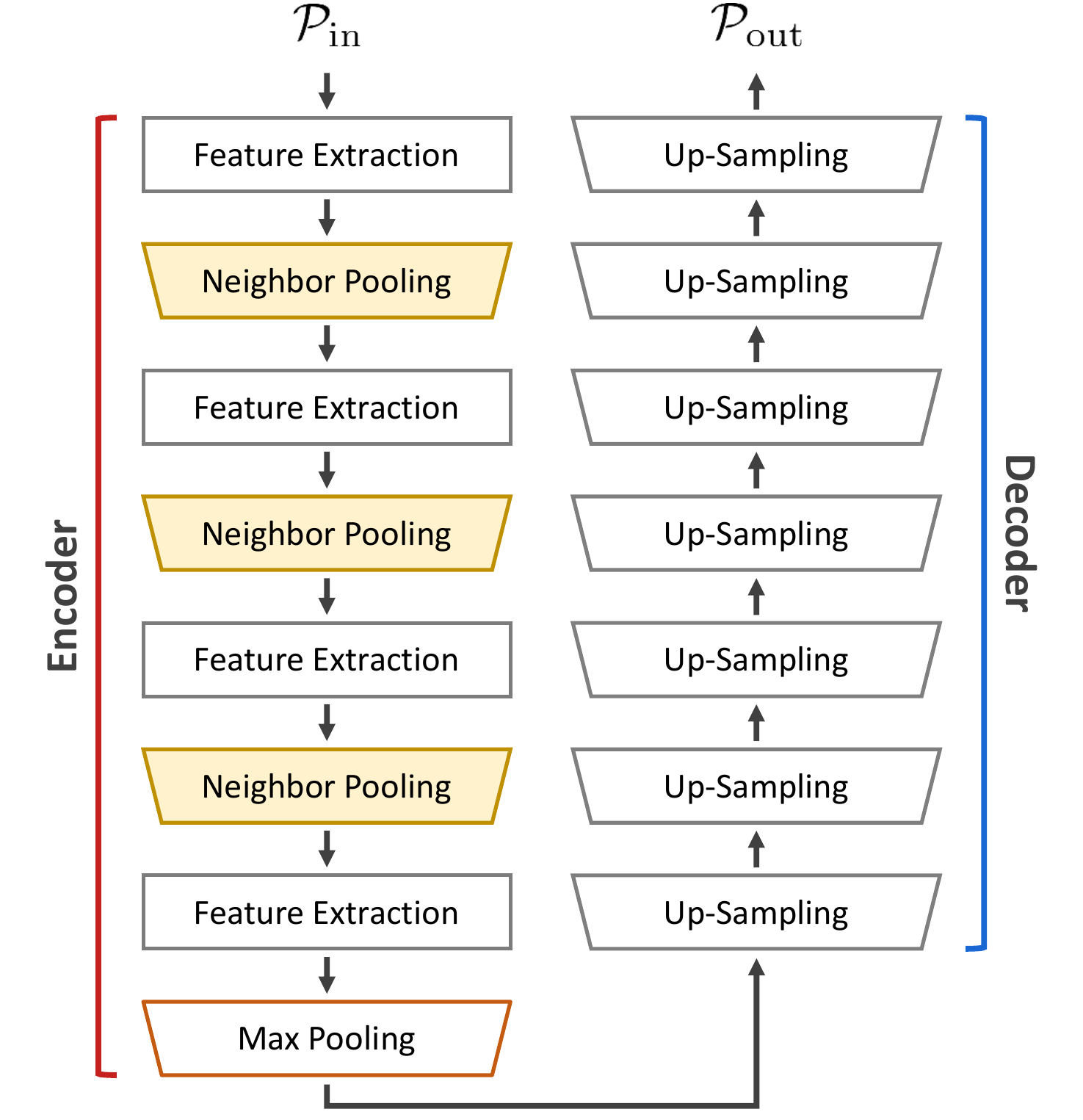}
\caption{This architecture is composed of the proposed operators to build its encoder and decoder.}
\label{fig:architecture_direct}
\end{figure}

\section{Encoder-decoder architectures}
\label{sec:architecture}

In order to uncover the strengths of our operators in \secref{sec:operation} (\ie feature extraction, neighbor pooling and up-sampling), we used them as building blocks to construct two different architectures. The first directly implements our operators to build an encoder-decoder while the second takes advantage of our operators to improve the transformers derived from PoinTr~\cite{yu2021pointr}.
We refer the readers to the Supplementary Materials for the detailed parameters of the architectures.

\subsection{Direct application}
\label{sec:direct}

The objective of the first architecture is to establish that building it solely from the proposed operators (with the additional max-pooling) can already be competitive in point cloud completion. 
We then propose an encoder-decoder architecture based on our operators alone as shown in \figref{fig:architecture_direct}.
The encoder is composed of four alternating layers of feature extraction and neighbor pooling. 
As the number of points from the input is reduced by 128 times, we use a max-pooling operator to extract a vector as our latent feature.
Taking the latent feature from the encoder, the decoder is then constructed from a series of up-sampling operators, resulting in a fine completion of 16,384 points.

\subsection{Transformers}

\begin{figure} [!t]
\centering
\includegraphics[width=0.85\linewidth]{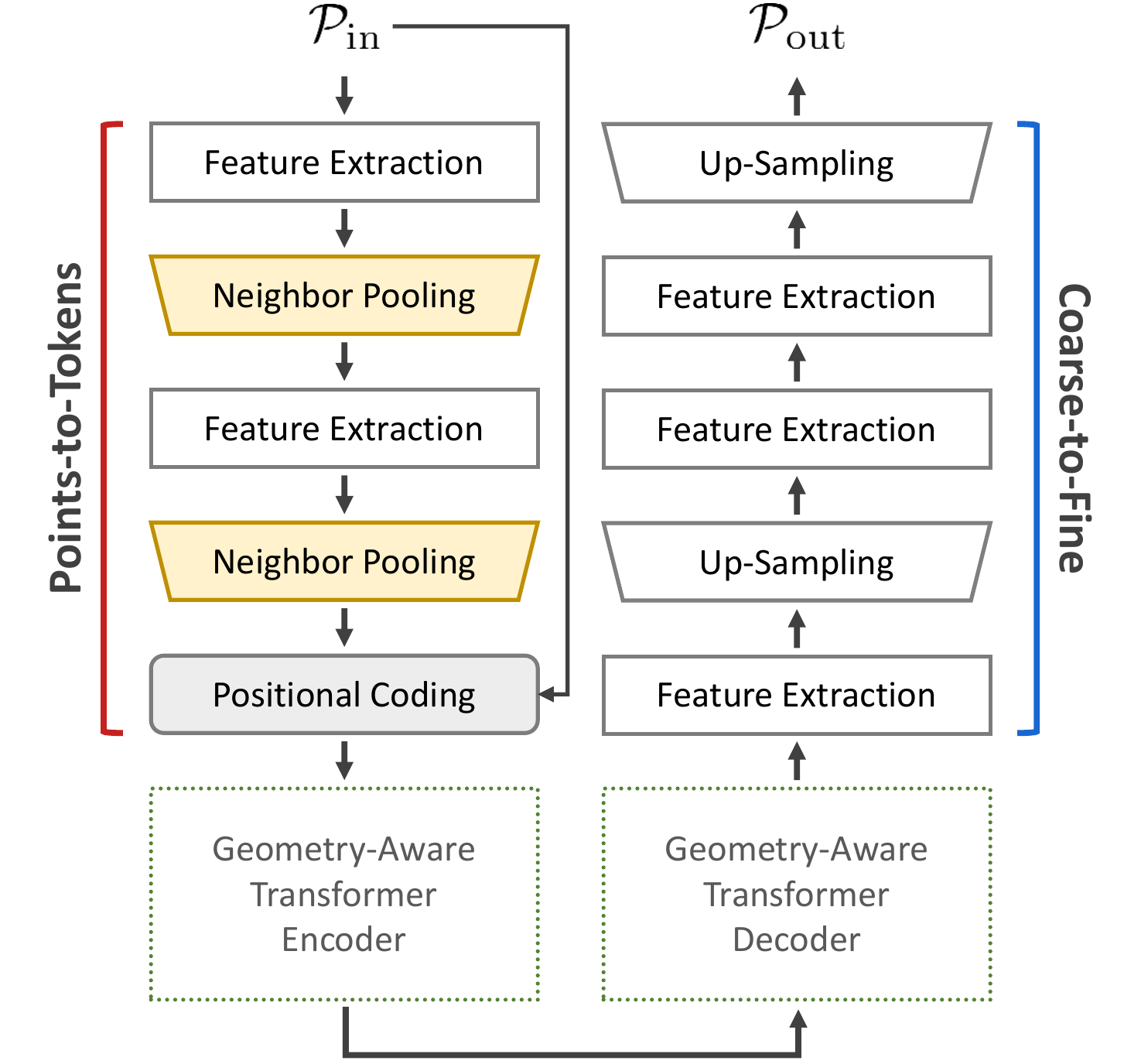}
\caption{This architecture is derived from the transformers backbone, where we use the proposed operators to convert the input 3D points to tokens and to perform the coarse-to-fine strategy.}
\label{fig:architecture_transformer}
\end{figure}

The second architecture aims at showing the diversity of the operators to improve the state-of-the-art from PoinTr~\cite{yu2021pointr} that uses transformers.
We therefore propose a transformer-based architecture that is derived from \cite{yu2021pointr} and our operators as summarized in \figref{fig:architecture_transformer}.

Before computing the attention mechanisms in the transformer, the partial scan are subsampled due to the memory constraint of the GPU. 
PoinTr~\cite{yu2021pointr} implements the Farthest Point Sampling (FPS) to reduce the number of points and MLP to convert the points to features. Conversely, our architecture applies the proposed operators. Similar to \secref{sec:direct}, this involves alternating the features extraction and neighbor pooling. 
Since the Fourier feature~\cite{tancik2020fourfeat} and SIRENs~\cite{sitzmann2020implicit} have proven that the sinusoidal activation is helpful in presenting complex signals and their derivatives in layer-by-layer structures, a positional coding based on the 3D coordinates is then added to the features. 
In \figref{fig:architecture_transformer}, we refer this block as \emph{points-to-token}.
Thereafter, we use the geometry-aware transformers from \cite{yu2021pointr} which produces a coarse point cloud. 

From the coarse point cloud, we then replace their coarse-to-fine strategy with our operators. This includes a series of alternating feature extraction and up-sampling operators as shown in \figref{fig:architecture_transformer}.

\begin{figure}[!t]
\centering
\includegraphics[width=1.0\linewidth]{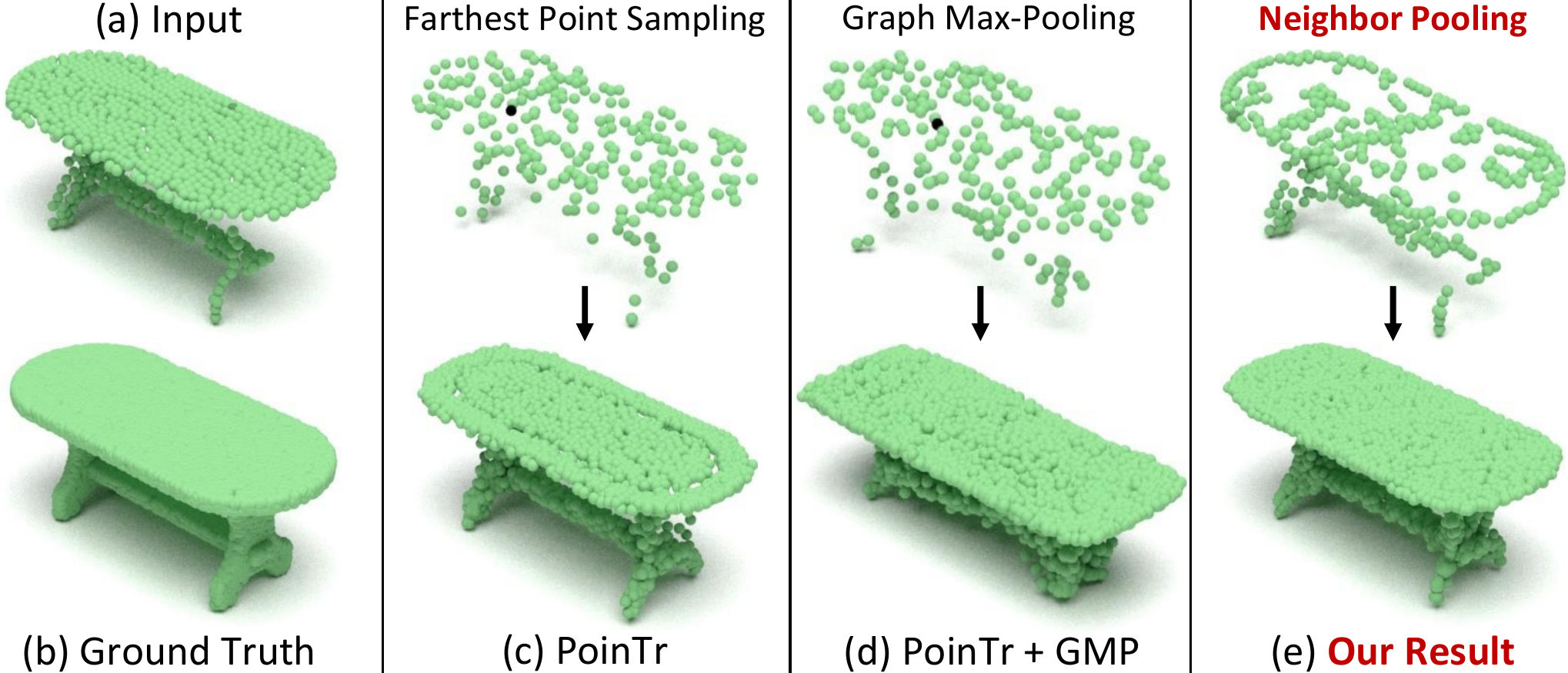}
\caption{The first row compares the point tokens chosen by
Farthest Point Sampling (FPS) in PointTr~\cite{yu2021pointr},
Graph Max-Pooling (GMP)~\cite{lin2020convolution} in PointTr~\cite{yu2021pointr}  and our proposed neighbor pooling in our transformer architecture. These tokens are then fed to the transformer and the  coarse-to-fine strategy to produce the reconstruction shown in the second row.  
}
\label{fig:neighborpool_querries}
\end{figure}

It it noteworthy to emphasize the difference between our architecture from PoinTr~\cite{yu2021pointr} and to understand the implication of the changes. 
The contributions of points-to-tokens and coarse-to-fine to the overall architecture is illustrated in \figref{fig:neighborpool_querries}. 
We can observe from this figure that the FPS from PoinTr~\cite{yu2021pointr} only finds the distant points while the results of our neighbor pooling sketches the contours of the input point cloud to capture the meaningful structures of the object. Notably, by looking at our sketch, we can already identify the that the object is a table. This is contrary to the random points from PoinTr~\cite{yu2021pointr}.
Moreover, our coarse-to-fine strategy uniformly reconstructs the planar region on the table as well as its base.
Later, in \secref{sec:ablation}, we numerically evaluate these advantages in order to show that the individual components has their own merits.

Since we previously discussed in \secref{sec:down} the difference of our down-sampling operation against 3D-GMP~\cite{lin2020convolution}, we became curious to see the reconstruction in \figref{fig:neighborpool_querries} if we replace the FPS in PoinTr~\cite{yu2021pointr} with the cosine similarity and GMP of \cite{lin2020convolution}. Similar to PoinTr, the new combination selects distant points as its tokens while the table in their final reconstruction increased in size.
In contrast, our tokens are more meaningful and the final results are more accurate.

\section{Loss functions}
\label{sec:loss}

Given the input point cloud $\Pin$ (\eg from a depth image), 
the objective of completion is to build the set of points $\Pout$ that fills up the missing regions in our input data.
Since we train our architecture in a supervised manner, we denote $\Pgt$ as the ground truth.

\vspace{-5pt}\paragraph{Completion.}
\label{sec:loss_completion}

To evaluate the predicted point cloud, we impose the Earth-moving distance~\cite{fan2017point}. 
Comparing the output points to the ground truth and vice-versa, we end up with
\begin{align}
\mathcal{L}_{\text{out}\rightarrow\text{gt}}
&= 
\sum_{p\in\Pout} \|p-\phi_\text{gt}(p)\|_2
\label{eq:l_out} \\
\mathcal{L}_{\text{gt}\rightarrow\text{out}}
&= 
\sum_{p\in\Pgt} \|p-\phi_\text{out}(p)\|_2
\label{eq:l_gt}
\end{align}
where $\phi_i(p)$ is a bijective function that finds the closest point in the point cloud $\cloud_i$ to $p$.

\vspace{-5pt}\paragraph{Order of points in $\Pout$.}

\begin{figure}[!t]
\centering
\includegraphics[width=0.85\linewidth]{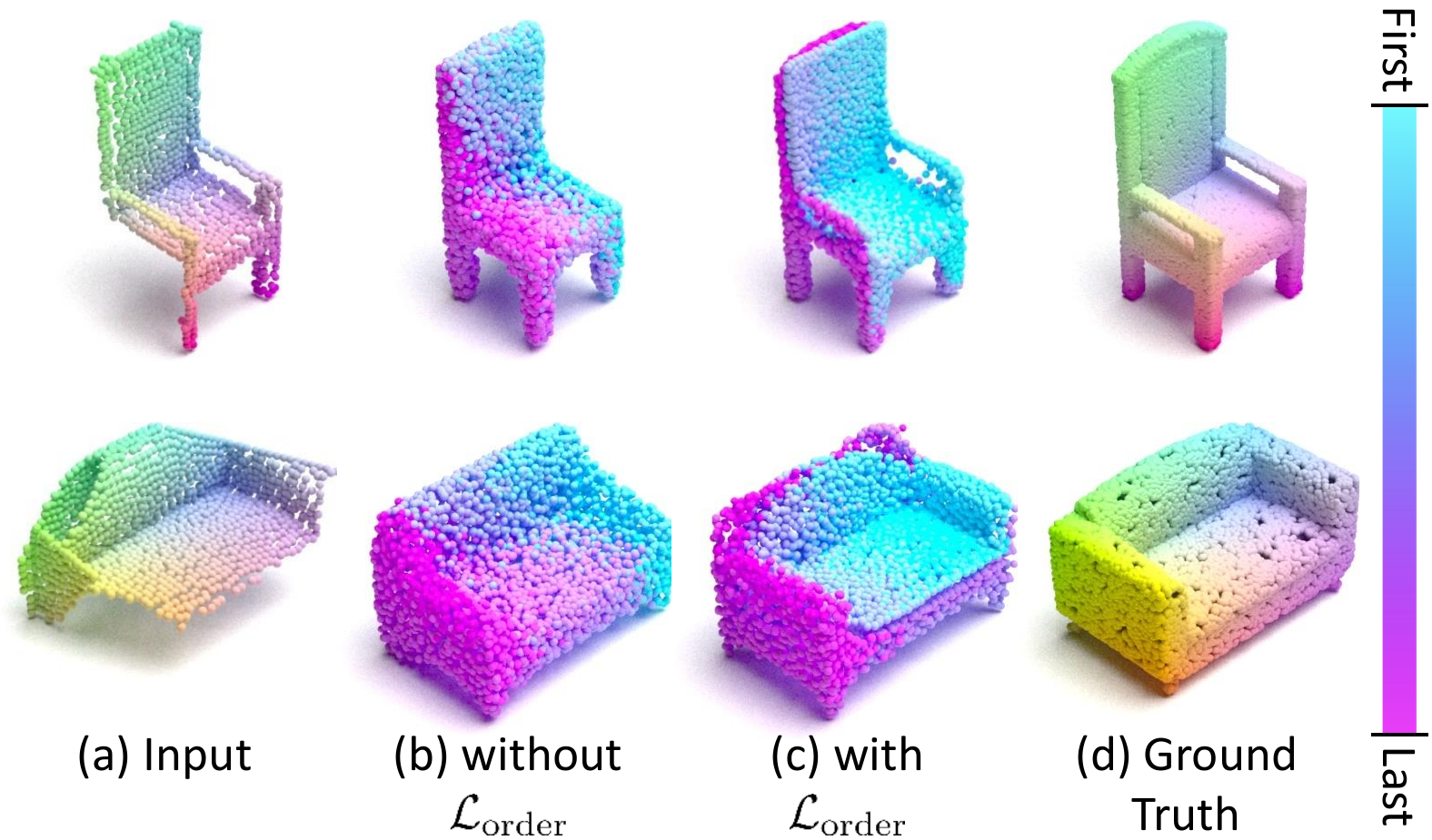}
\caption{Compares the order of the point clouds reconstructed in the object completion with and without $\mathcal{L}_\text{order}$}
\label{fig:complete_observe}
\end{figure}

After training with \eqref{eq:l_out} and \eqref{eq:l_gt}, we noticed that the points in the output reconstruction are ordered from left to right as shown in \figref{fig:complete_observe}(b).
We want to take advantage of this organization and investigate this behavior further.
Assuming the idea that, among the points in $\Pout$, we are confident that the input point cloud must be part of it, we introduce a loss function that enforces that the first subset in $\Pout$ is similar to $\Pin$. 
We formally write this loss function as
\begin{align}
\mathcal{L}_\text{order}
&= 
\sum_{p\in\Pin} \mathcal{S}(\theta_\text{out}(p)) \cdot \|p-\phi_\text{out}(p)\|_2
\label{eq:l_order}
\end{align}
where $\theta_\text{out}(p)$ is the index of the closest point in $\Pout$ based on $\phi_\text{out}(p)$ while 
\begin{align}
\mathcal{S}(\theta) =
\begin{cases}
    1, & \text{if } \theta\leq |\Pin|\\
    0, & \text{otherwise}
\end{cases}
\label{eq:step}
\end{align}
is a step function that returns one if the index is within the first $|\Pin|$ points.

When we plot the results with $\mathcal{L}_\text{order}$ in \figref{fig:complete_observe}(c), we noticed that the order in $\Pout$ moves from the observed to the occluded. In addition, fine-grained geometrical details such as the armrest of the chair are visible when training with $\mathcal{L}_\text{order}$; thus, improving the overall reconstruction.

\begin{figure*}[!t]
\centering
\includegraphics[width=1.0\linewidth]{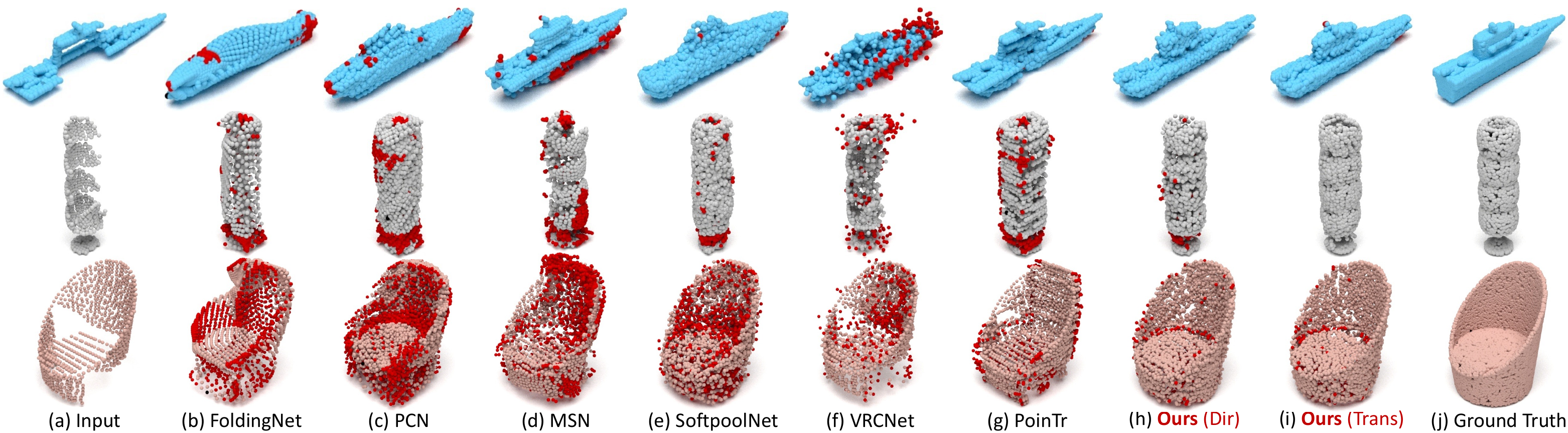}
\caption{Object completion results where we highlight the errors in red points. 
}
\label{fig:shapenet_qualitatives}
\end{figure*}

\vspace{-5pt}\paragraph{Semantic scene completion.}

In addition to the architecture in \secref{sec:architecture} and the loss functions in \eqref{eq:l_out}, \eqref{eq:l_gt} and \eqref{eq:l_order} for completion, a semantic label is added to each point in the predicted cloud $\Pout$. 
Given $N_c$ categories, we denote the label for each point as a one-hot code $\mathbf{l}_i= [l_{i,c}]_{c=1}^{n_c}$ for the $i$-th point in $\Pout$ and the $c$-th category. 
Since training is supervised, the ground truth point clouds are also labeled with the semantic category.

After establishing the correspondence between the predicted point cloud to the ground truth in \eqref{eq:l_out} in training, we also extract the ground truth semantic label $\hat{\mathbf{l}}_i$.
It then follows that the binary cross-entropy of the $i$-th point is computed 
\begin{align}
    \epsilon_i = 
    - \frac{1}{N_c} \sum_{c=i}^{N_s}
    \hat{l}_{i,c} \log l_{i,c} + 
    (1 - \hat{l}_{i,c}) ( 1 - \log l_{i,c} ) 
\end{align}
and formulate the semantic loss function as 
\begin{align}
    \mathcal{L}_\text{semantic} = \frac{\gamma}{|\mathcal{P}_\text{in}|}\sum_{i=i}^{|\mathcal{P}_\text{in}|} \epsilon_i
\end{align}
where the weight
\begin{align}
    \gamma = 
    \frac{0.01}{\mathcal{L}_{\text{out}\rightarrow\text{gt}} 
    + \mathcal{L}_{\text{gt}\rightarrow\text{out}}}
    \label{eq:semantic_weight}
\end{align}
triggers to increase the influence of the $\mathcal{L}_\text{semantic}$ in training as the completion starts to converge. 
Note that $\gamma$ is an important factor, since the output point cloud is erratic in the initial iterations, which means that it can abruptly change from one iteration to the next before the completion starts converging.

\section{Experiments}

To highlight the strengths of the proposed method, this section focuses on two experiments -- object completion and semantic scene completion.

\subsection{Object completion}

We evaluate the geometric completion of a single object on the ShapeNet~\cite{chang2015shapenet} database where they have the point clouds of the partial scans as input and their corresponding ground truth completed shape. 
The input scans are composed of 2,048 points while the database provides a low resolution output of 2,048 points and a high resolution of 16,384 points. 
We follow the standard evaluation on 8 categories where all objects are roughly normalized into the same scale with point coordinates ranging between $-1$ to $1$.

\vspace{-5pt}\paragraph{Numerical results.}

We conduct our experiments based on three evaluation strategies from Completion3D~\cite{tchapmi2019topnet}, PCN~\cite{yuan2018pcn} and MVP~\cite{pan2021variational}. 
Evaluating on 8 objects 
(\textit{plane}, \textit{cabinet}, \textit{car}, \textit{chair}, \textit{lamp}, \textit{sofa}, \textit{table}, \textit{vessel}), 
they measure the predicted reconstruction through the L2-Chamfer distance, L1-Chamfer distance and the F-Score@1\%, respectively. 
Note that, in this paper, we also follow the standard protocol where the value presented for the Chamfer distance is multiplied by $10^3$.
Although \tabref{tab:shapenet} only shows the average results across all categories, we refer the readers to the supplementary materials for the more detailed comparison.

One of the key observations in this table is the capacity of our direct architecture to surpass most of the other methods' results. Among 11 approaches, our Chamfer distance is only worse than 3 methods while our F-Score@1\% is better than all of them. This therefore establishes the strength of our operators since our first architecture is solely composed of it.
Moreover, our second architecture, which combines our operators with the transformer, reduces the error by 3-5\% on the Chamfer distance and increases the accuracy by 4.5\% on the F-Score@1\%. 

The table also examines the effects of  $\mathcal{L}_\text{order}$ to our reconstruction. 
Training with $\mathcal{L}_\text{order}$ improves our results by 0.12-0.13 in Chamfer distance and 0.013-0.021 in F-Score@1\%, validating our observations in \figref{fig:complete_observe}.

\begin{table}[!t]
\centering
\resizebox{\linewidth}{!}
{
\begin{tabular}{l|ccc}
\toprule	
 \multicolumn{1}{c}{} 
 & \small{Completion3D} & \small{PCN} & \small{MVP} \\
 \multicolumn{1}{c}{Method} 
 & \small\emph{L2-Chamfer} & \small\emph{L1-Chamfer} & \small\emph{F-Score@1\%} \\
\midrule 
       FoldingNet~\cite{yang2018foldingnet} & 19.07 & 14.31 & -- \\ 
       SoftPoolNet~\cite{wang_softpool} & 11.07 & 9.20 & 0.666 \\ 
       TopNet~\cite{tchapmi2019topnet} & 14.25 & 12.15 & 0.576 \\ 
       PCN~\cite{yuan2018pcn} & 18.22 & 9.64 & 0.614 \\
       MSN~\cite{liu2020morphing} & -- & 9.97 & 0.690 \\
       GRNet~\cite{grnet_xie} & 10.64 & 8.83 & 0.677 \\
       ECG~\cite{pan2020ecg} &  -- & --  & 0.736 \\
       NSFA~\cite{zhang2020detail} &  -- & --  & 0.770 \\
       CRN~\cite{Wang_2020_CVPR} & 9.21 & 8.51  & 0.724 \\
       SCRN~\cite{wang2020self} & 9.13 & 8.29 & -- \\
       VRCNet~\cite{pan2021variational} & 8.12 & --  & 0.781 \\
       PoinTr~\cite{yu2021pointr} & 9.22  & 8.38 & 0.741 \\
       ASFM-Net~\cite{xia2021asfm} & 6.68 & -- & -- \\
\midrule 
       \textbf{Ours} \small{(Direct)} & 8.35 & 8.46 & 0.801 \\
       \small{--\textit{without} $\mathcal{L}_\text{order}$} & 8.47 & 8.59 & 0.788 \\
       \small{--\textit{input} $\mathcal{P}_\text{gt}$} & 5.11 & 5.37 & 0.923 \\
\midrule 
       \textbf{Ours} \small{(Transformer)} & \textbf{6.64} & \textbf{7.96} & \textbf{0.816} \\
       \small{--\textit{without} $\mathcal{L}_\text{order}$} & 6.74 & 8.09 & 0.795 \\
       \small{--\textit{input} $\mathcal{P}_\text{gt}$} & 4.46 & 4.95 & 0.962 \\
\bottomrule
\end{tabular}
}
\caption{Evaluation on Completion3D~\cite{tchapmi2019topnet}, PCN~\cite{yuan2018pcn} and MVP~\cite{pan2021variational} datasets with their corresponding metrics for the object completion task.
 \label{tab:shapenet}
}
\end{table}

\vspace{-5pt}\paragraph{Qualitative results.}

We compare our object completion results  in \figref{fig:shapenet_qualitatives} with the recently proposed methods: 
FoldingNet~\cite{yang2018foldingnet}, 
PCN~\cite{yuan2018pcn}, 
MSN~\cite{liu2020morphing}, 
SoftPoolNet~\cite{wang_softpool},
VRCNet~\cite{pan2021variational} and
PoinTr~\cite{yu2021pointr}.
The red points in the figure highlight the errors in the reconstruction. 
All the approaches reconstructs a point cloud with 16,384 points with the exception for FoldingNet with 2,048 points and MSN with 8,192. 

Since FoldingNet and PCN take advantage of their mathematical assumption where they rely on deforming one or more planar grids, they tend to over-smooth their reconstruction where finer details such as the boat is flattened. In contrast, our method can perform better on the smooth regions as well as the finer structures.  
Nevertheless, the more recent approaches like \cite{liu2020morphing,wang_softpool,pan2021variational,yu2021pointr} can also produce more descriptive reconstruction on the boat. However, they produce more errors which is highlighted in the unconventional lamp or chair. Overall, our reconstructions are closer to the ground truth.

\vspace{-5pt}\paragraph{Failure cases.}

\begin{figure}[!b]
\centering
\includegraphics[width=1.0\linewidth]{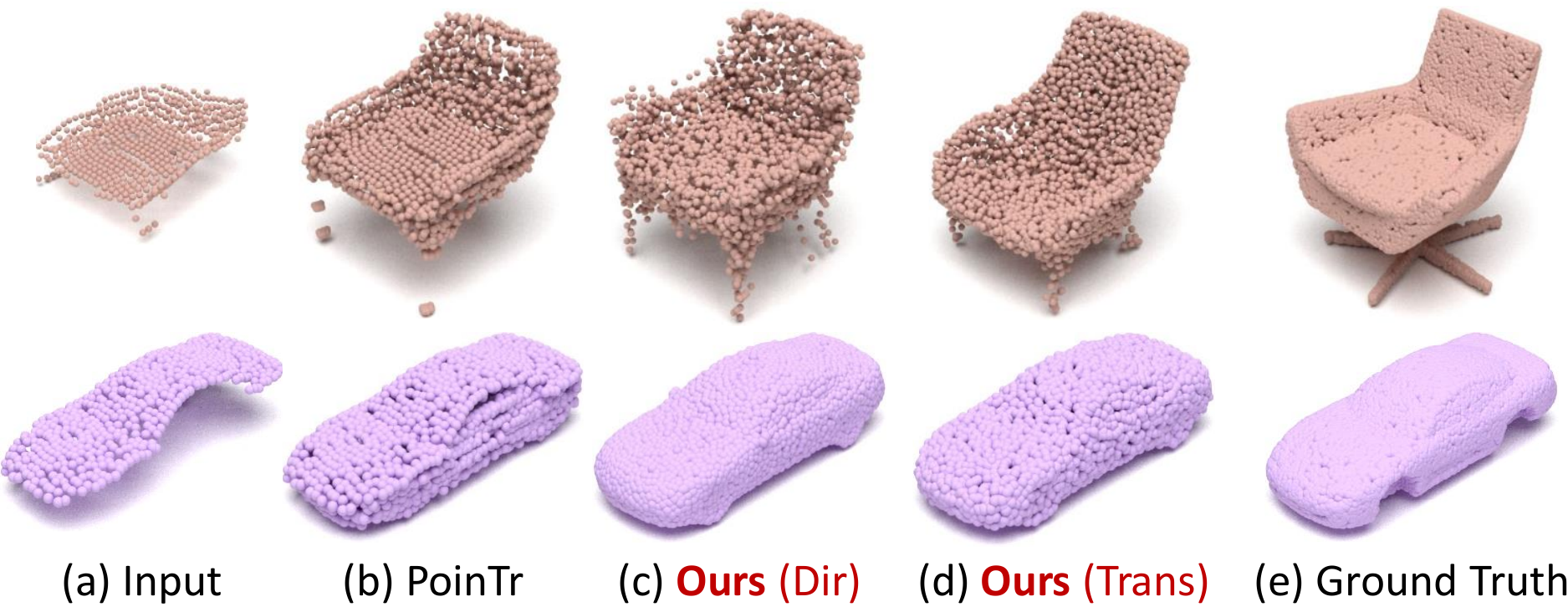}
\caption{Examples of the failure cases in object completion.}
\label{fig:failures}
\end{figure}

In addition to the qualitative results, we also examine the failure cases in \figref{fig:failures}. Most of them are objects with unusual structures like the car without the wheels. Another issue is when there is an insufficient amount of input point cloud to describe the object such as the chair.
Notably, compared to the state-of-the-art, our reconstructions are still better in these situations.

\subsection{Semantic scene completion}

This evaluation aims at reconstructing the scene from a single depth image through a point cloud or an SDF volume where each point or voxel is categorized with a semantic class.
Originally introduced for 2.5D semantic segmentation, NYU~\cite{silberman2012indoor} and ScanNet~\cite{dai2017scannet}, which were later annotated for semantic completion by~\cite{song2017semantic, wu2020scfusion}, are among the most relevant benchmark datasets in this field. These datasets include pairs of depth image and the corresponding semantically labeled 3D reconstruction.

\begin{table}[!t]
\centering
\resizebox{0.8\linewidth}{!}
{
\begin{tabular}{l|c|c}
\toprule
 \multicolumn{1}{c}{Method} 
 & Resolution &  \emph{Average IoU} \\
\midrule 
    Lin \etal~\cite{lin2013holistic} & 60 & 12.0 \\
    Geiger and Wang~\cite{geiger2015joint} & 60 & 19.6 \\
    SSCNet~\cite{song2017semantic} & 60 & 30.5 \\
    VVNet~\cite{guo2018view} & 60 & 32.9 \\
    SaTNet~\cite{liu2018nips} & 60 & 34.4 \\
    ForkNet~\cite{wang2019forknet} & 80 & 37.1 \\
    CCPNet~\cite{zhang2019cascaded} & 240 & 38.5 \\
    SketchSSC~\cite{chen20203d} & 60 & 41.1 \\
    SISNet~\cite{cai2021semantic} & 60 & \textbf{52.4} \\
\midrule 
    \textbf{Ours} \small{(Direct)} & 60 & 40.0 \\
    \small{--\textit{with} $\gamma=1$ \textit{in} $\mathcal{L}_\text{semantic}$} & 60 & 37.2 \\
\midrule 
    \textbf{Ours} \small{(Transformer)} & 60 & 42.4 \\ 
    \small{--\textit{with} $\gamma=1$ \textit{in} $\mathcal{L}_\text{semantic}$} & 60 & 38.9 \\
\bottomrule
\end{tabular}
}
\caption{Semantic scene completion on NYU~\cite{silberman2012indoor} dataset. The value in resolution~($x$) is the output volumetric resolution which is $x \times 0.6x \times x$.
 \label{tab:nyu}
}
\end{table}

\vspace{-5pt}\paragraph{Semantic scene completion with voxels.}

NYU are provided with real scans for indoor scenes which are acquired with a Kinect depth sensor.
Following SSCNet~\cite{song2017semantic}, the semantic categories include 12 classes of varying shapes and sizes: \emph{empty space}, \emph{ceiling}, \emph{floor}, \emph{wall}, \emph{window}, \emph{chair}, \emph{bed}, \emph{sofa}, \emph{table}, \emph{tvs}, \emph{furniture} and \emph{other objects}.

Since the other point cloud completion do not handle semantic segmentation, we start our evaluation by comparing with the voxel-based approaches which perform the both the completion and the semantic segmentation such as 
\cite{lin2013holistic,geiger2015joint,song2017semantic,guo2018view,liu2018nips,wang2019forknet,zhang2019cascaded,chen20203d,cai2021semantic}.
Considering that the volumetric data evaluates through the IoU, we need to convert our point clouds to voxel grids to make the comparison. 

One of the significant advantage of point clouds over voxels is that we are not constrained to a specific resolution. Since most method evaluate on $60 \times 36 \times 60$, we converted our point cloud to this resolution.
Our approach achieves competitive average IoU of 42.4\% 
which is better than all the other methods except for SISNet~\cite{cai2021semantic}.
%
However, it is noteworthy to mention that our method faces additional errors associated to the conversion from point cloud to voxels. 
In addition, the ground truth voxels for the furnitures in the NYU dataset is a solid volume which is not a plausible format for point cloud approaches which focuses more on the surface reconstruction.
%
This in effect decreases the IoU of our method.

Moreover, \tabref{tab:nyu} includes a small ablation study to verify the contribution of $\gamma$ from \eqref{eq:semantic_weight} in $\mathcal{L}_\text{semantic}$. 
If we discard \eqref{eq:semantic_weight} by setting $\gamma$ to one, the IoU for our models decrease by 7.5-9\%; thus, proving the advantage in adaptively weighing the semantic loss function.

\begin{table}[!t]
\centering
\resizebox{0.78\linewidth}{!}
{
\begin{tabular}{l|ccc}
\toprule	
 \multicolumn{1}{c}{Method} 
 & ~~CompleteScanNet~~ & ~~NYU~~ \\
\midrule 
       FoldingNet~\cite{yang2018foldingnet} & 11.25 & 14.66 \\
       AtlasNet~\cite{Groueix_2018_CVPR} & 8.92 & 10.12 \\
       PCN~\cite{yuan2018pcn} & 8.19 & 9.98 & \\
       MSN~\cite{liu2020morphing} & 7.28 & 8.65 \\
       SoftPoolNet~\cite{wang_softpool} & 8.27 & 9.29 \\
       GRNet~\cite{grnet_xie} & 4.56 & 5.80 \\
       VRCNet~\cite{pan2021variational} & 4.29 & 5.45 \\
       PoinTr~\cite{yu2021pointr} & 5.08 & 5.92 \\
\midrule 
       \textbf{Ours} \small{(Direct)} & 3.17 & 4.72 \\
       \textbf{Ours} \small{(Transformer)} & \textbf{3.04} & \textbf{4.38} \\
\bottomrule
\end{tabular}
}
\caption{Evaluation on CompleteScanNet~\cite{wu2020scfusion} and NYU~\cite{silberman2012indoor} dataset for scene completion, measuring the average Chamfer distance trained with L2 distance (multiplied by $10^3$) with the output resolution of 16,384.
\label{tab:scannet}
}
\end{table}

\vspace{-5pt}\paragraph{Point cloud scene completion.}

Another relevant dataset is from ScanNet~\cite{dai2017scannet} which was supplemented with the ground truth semantic completion by CompleteScanNet~\cite{wu2020scfusion}.
This include a total of 45,451 paired partial scan and semantic completion for training.
Our evaluation in \tabref{tab:scannet} takes 2,048 points as input and reconstructs the scene with 16,384 points.
Since there is no previous work that focused on point cloud scene completion, we compare against methods that were designed for a single object completion such as PCN~\cite{yuan2018pcn}, MSN~\cite{liu2020morphing}, SoftPoolNet~\cite{wang_softpool} and GRNet~\cite{grnet_xie}.
Based on our evaluation in \tabref{tab:scannet}, both versions of our architectures attain the best results. 
Notably, we also compared these methods on the NYU dataset in \tabref{tab:scannet}. Similarly, the proposed architectures also achieve the  state-of-the-art in point cloud scene completion.

\newcommand{\high}[1]{\cellcolor[HTML]{FFF2CC}{#1}}

\begin{table}[!b]
\centering
\resizebox{0.85\linewidth}{!}
{
\begin{tabular}{l|ccc|c}
\multicolumn{5}{c}{\textsc{object completion}}\\
\toprule
 \multicolumn{1}{c}{} 
 & \multicolumn{4}{c}{\small{Coarse-to-Fine}}  \\
 \cmidrule(lr){2-5}
 \multicolumn{1}{c}{\small{Backbone}} 
 & deform & deconv & \multicolumn{1}{c}{EFE} & \multicolumn{1}{c}{\textbf{Ours}} \\
\midrule 
      MSN~\cite{liu2020morphing} & \high{7.28} & 9.34 & 7.15 & 6.91 \\
      PoinTr~\cite{yu2021pointr} & \high{5.48} & 5.71 & 4.91 & 3.76 \\
      SoftPoolNet~\cite{wang_softpool} & 10.08 & \high{8.27} & 7.65 & 7.63 \\
      GRNet~\cite{grnet_xie} & 9.25 & \high{5.61} & 5.26 & 4.90 \\
      VRCNet~\cite{pan2021variational} & 8.09 & 8.88 & \high{5.08} & 4.21 \\
\midrule 
       
      \textbf{Ours} & 4.93 & 4.99 & 4.12 & \high{\textbf{3.04}} \\
\bottomrule
\multicolumn{5}{c}{}\\
\multicolumn{5}{c}{\textsc{scene completion}}\\
\toprule
 \multicolumn{1}{c}{} 
 & \multicolumn{4}{c}{\small{Coarse-to-Fine}}  \\
 \cmidrule(lr){2-5}
 \multicolumn{1}{c}{\small{Backbone}} 
 & deform & deconv & \multicolumn{1}{c}{EFE} & \multicolumn{1}{c}{\textbf{Ours}} \\
\midrule 
      MSN~\cite{liu2020morphing} & \high{9.97} & 12.31 & 9.26 & 9.08 \\
      PoinTr~\cite{yu2021pointr} & \high{8.38} & 8.49 & 8.31 & 8.13 \\
      TreeGAN~\cite{treegan2019} & 14.26 & \high{9.72} & 9.12 & 9.05 \\
      SoftPoolNet~\cite{wang_softpool} & 11.73 & \high{9.20} & 8.75 & 8.64 \\
      GRNet~\cite{grnet_xie} & 9.12 & \high{8.83} & 8.73 & 8.51 \\
      VRCNet~\cite{pan2021variational} & 10.03 & 10.20 & \high{8.52} & 8.26 \\
\midrule 
      \textbf{Ours} & 8.19 & 8.30 & 8.07 & \high{\textbf{7.96}} \\
\bottomrule
\end{tabular}
}
\caption{Mix-and-match evaluation on different backbone attached to different coarse-to-fine methods for object and scene completion.
The originally proposed combinations are marked in yellow.
\label{tab:refiner}
}
\end{table}

\section{Ablation study}
\label{sec:ablation}

This section focuses on the strengths of our operator in our transformer architecture. Although we adapt the transformer from PoinTr~\cite{yu2021pointr}, we argue that every component we added is significant to the overall performance.
To evaluate this, we disentangle the points-to-tokens and coarse-to-fine blocks. 
In practice, we separate the backbone, which takes points in the partial scan as input and outputs a coarse point cloud, from the coarse-to-fine strategy.
Evidently, in our approach, the points-to-tokens block is part of the backbone.

Since most methods can also be separated in this manner, we then compose \tabref{tab:refiner} to mix-and-match different backbones with different coarse-to-fine methods for object and scene completion.
In both tables, we classified the other coarse-to-fine methods as: 
(1)~\emph{deform} which includes the operation in deforming 3D grids; 
(2)~\emph{deconv} which processes with MLP, 1D or 2D deconvolutions; 
and, 
(3)~Edge-aware Feature Expansion (\emph{EFE}) \cite{pan2020ecg}.
We then highlight the originally proposed architectures in yellow.

For any given backbone in every row, our coarse-to-fine method produces the best results. Moreover, for any given coarse-to-fine strategy in every column, our backbone performs the best.
Therefore, this study essentially proves that each of the proposed components in our transformer architecture has a significant role in the overall performance.

\section{Conclusion}

We propose three novel operators for point cloud processing. 
To bring out the value of these operators, we apply them on two novel architectures that are designed for object completion and semantic scene completion. 
The first assembles together the proposed operators in an encoder-decoder fashion, while the second incorporates them in the context of transformers.
Notably, both architectures produce highly competitive results, with the latter achieving the state of the art in point cloud completion for both objects and scenes.

{\small
\bibliographystyle{ieee_fullname}
\bibliography{PaperForReview}

\begin{thebibliography}{10}\itemsep=-1pt

\bibitem{besl1992method}
Paul~J Besl and Neil~D McKay.
\newblock Method for registration of 3-d shapes.
\newblock In {\em Sensor fusion IV: control paradigms and data structures},
  volume 1611, pages 586--606. International Society for Optics and Photonics,
  1992.

\bibitem{cai2021semantic}
Yingjie Cai, Xuesong Chen, Chao Zhang, Kwan-Yee Lin, Xiaogang Wang, and
  Hongsheng Li.
\newblock Semantic scene completion via integrating instances and scene
  in-the-loop.
\newblock In {\em Proceedings of the IEEE/CVF Conference on Computer Vision and
  Pattern Recognition}, pages 324--333, 2021.

\bibitem{chang2015shapenet}
Angel~X Chang, Thomas Funkhouser, Leonidas Guibas, Pat Hanrahan, Qixing Huang,
  Zimo Li, Silvio Savarese, Manolis Savva, Shuran Song, Hao Su, et~al.
\newblock Shapenet: An information-rich 3d model repository.
\newblock {\em arXiv preprint arXiv:1512.03012}, 2015.

\bibitem{chen20203d}
Xiaokang Chen, Kwan-Yee Lin, Chen Qian, Gang Zeng, and Hongsheng Li.
\newblock 3d sketch-aware semantic scene completion via semi-supervised
  structure prior.
\newblock In {\em Proceedings of the IEEE/CVF Conference on Computer Vision and
  Pattern Recognition}, pages 4193--4202, 2020.

\bibitem{chen1992object}
Yang Chen and G{\'e}rard Medioni.
\newblock Object modelling by registration of multiple range images.
\newblock {\em Image and vision computing}, 10(3):145--155, 1992.

\bibitem{dai2017scannet}
Angela Dai, Angel~X Chang, Manolis Savva, Maciej Halber, Thomas Funkhouser, and
  Matthias Nie{\ss}ner.
\newblock Scannet: Richly-annotated 3d reconstructions of indoor scenes.
\newblock In {\em Proceedings of the IEEE Conference on Computer Vision and
  Pattern Recognition}, pages 5828--5839, 2017.

\bibitem{dai2017shape}
Angela Dai, Charles~Ruizhongtai Qi, and Matthias Nie{\ss}ner.
\newblock Shape completion using 3d-encoder-predictor cnns and shape synthesis.
\newblock In {\em Proc. IEEE Conf. on Computer Vision and Pattern Recognition
  (CVPR)}, volume~3, 2017.

\bibitem{dai2018scancomplete}
Angela Dai, Daniel Ritchie, Martin Bokeloh, Scott Reed, J{\"u}rgen Sturm, and
  Matthias Nie{\ss}ner.
\newblock Scancomplete: Large-scale scene completion and semantic segmentation
  for 3d scans.
\newblock In {\em Proceedings of the IEEE Conference on Computer Vision and
  Pattern Recognition}, pages 4578--4587, 2018.

\bibitem{fan2017point}
Haoqiang Fan, Hao Su, and Leonidas~J Guibas.
\newblock A point set generation network for 3d object reconstruction from a
  single image.
\newblock In {\em Proceedings of the IEEE conference on computer vision and
  pattern recognition}, pages 605--613, 2017.

\bibitem{geiger2015joint}
Andreas Geiger and Chaohui Wang.
\newblock Joint 3d object and layout inference from a single rgb-d image.
\newblock In {\em German Conference on Pattern Recognition}, pages 183--195.
  Springer, 2015.

\bibitem{Groueix_2018_CVPR}
Thibault Groueix, Matthew Fisher, Vladimir~G. Kim, Bryan~C. Russell, and
  Mathieu Aubry.
\newblock A papier-mâché approach to learning 3d surface generation.
\newblock In {\em The IEEE Conference on Computer Vision and Pattern
  Recognition (CVPR)}, June 2018.

\bibitem{guo2018view}
Yuxiao Guo and Xin Tong.
\newblock View-volume network for semantic scene completion from a single depth
  image.
\newblock In {\em Proceedings of the International Joint Conference on
  Artificial Intelligence (IJCAI)}. AAAI Press, 2018.

\bibitem{li2018pointcnn}
Yangyan Li, Rui Bu, Mingchao Sun, Wei Wu, Xinhan Di, and Baoquan Chen.
\newblock Pointcnn: Convolution on x-transformed points.
\newblock In {\em Advances in Neural Information Processing Systems}, pages
  820--830, 2018.

\bibitem{lin2013holistic}
Dahua Lin, Sanja Fidler, and Raquel Urtasun.
\newblock Holistic scene understanding for 3d object detection with rgbd
  cameras.
\newblock In {\em Proceedings of the IEEE international conference on computer
  vision}, pages 1417--1424, 2013.

\bibitem{lin2020convolution}
Zhi-Hao Lin, Sheng-Yu Huang, and Yu-Chiang~Frank Wang.
\newblock Convolution in the cloud: Learning deformable kernels in 3d graph
  convolution networks for point cloud analysis.
\newblock In {\em Proceedings of the IEEE/CVF Conference on Computer Vision and
  Pattern Recognition}, pages 1800--1809, 2020.

\bibitem{liu2020morphing}
Minghua Liu, Lu Sheng, Sheng Yang, Jing Shao, and Shi-Min Hu.
\newblock Morphing and sampling network for dense point cloud completion.
\newblock In {\em Proceedings of the AAAI Conference on Artificial
  Intelligence}, volume~34, pages 11596--11603, 2020.

\bibitem{liu2018nips}
Shice Liu, YU HU, Yiming Zeng, Qiankun Tang, Beibei Jin, Yinhe Han, and Xiaowei
  Li.
\newblock See and think: Disentangling semantic scene completion.
\newblock In S. Bengio, H. Wallach, H. Larochelle, K. Grauman, N. Cesa-Bianchi,
  and R. Garnett, editors, {\em Advances in Neural Information Processing
  Systems 31}, pages 263--274. Curran Associates, Inc., 2018.

\bibitem{occupancy_Mescheder}
Lars Mescheder, Michael Oechsle, Michael Niemeyer, Sebastian Nowozin, and
  Andreas Geiger.
\newblock Occupancy networks: Learning 3d reconstruction in function space.
\newblock In {\em Proceedings IEEE Conf. on Computer Vision and Pattern
  Recognition (CVPR)}, 2019.

\bibitem{pan2020ecg}
Liang Pan.
\newblock Ecg: Edge-aware point cloud completion with graph convolution.
\newblock {\em IEEE Robotics and Automation Letters}, 5(3):4392--4398, 2020.

\bibitem{pan2021variational}
Liang Pan, Xinyi Chen, Zhongang Cai, Junzhe Zhang, Haiyu Zhao, Shuai Yi, and
  Ziwei Liu.
\newblock Variational relational point completion network.
\newblock In {\em Proceedings of the IEEE/CVF Conference on Computer Vision and
  Pattern Recognition}, pages 8524--8533, 2021.

\bibitem{park2019deepsdf}
Jeong~Joon Park, Peter Florence, Julian Straub, Richard Newcombe, and Steven
  Lovegrove.
\newblock Deepsdf: Learning continuous signed distance functions for shape
  representation.
\newblock In {\em Proceedings of the IEEE Conference on Computer Vision and
  Pattern Recognition}, pages 165--174, 2019.

\bibitem{qi2017pointnet}
Charles~R Qi, Hao Su, Kaichun Mo, and Leonidas~J Guibas.
\newblock Pointnet: Deep learning on point sets for 3d classification and
  segmentation.
\newblock In {\em Proceedings of the IEEE Conference on Computer Vision and
  Pattern Recognition}, pages 652--660, 2017.

\bibitem{qi2017pointnet++}
Charles~Ruizhongtai Qi, Li Yi, Hao Su, and Leonidas~J Guibas.
\newblock Pointnet++: Deep hierarchical feature learning on point sets in a
  metric space.
\newblock In {\em Advances in Neural Information Processing Systems (NIPS)},
  2017.

\bibitem{treegan2019}
Dong~Wook Shu, Sung~Woo Park, and Junseok Kwon.
\newblock 3d point cloud generative adversarial network based on tree
  structured graph convolutions.
\newblock In {\em Proceedings of the IEEE/CVF International Conference on
  Computer Vision}, pages 3859--3868, 2019.

\bibitem{silberman2012indoor}
Nathan Silberman, Derek Hoiem, Pushmeet Kohli, and Rob Fergus.
\newblock Indoor segmentation and support inference from rgbd images.
\newblock In {\em European conference on computer vision}, pages 746--760.
  Springer, 2012.

\bibitem{sitzmann2020implicit}
Vincent Sitzmann, Julien Martel, Alexander Bergman, David Lindell, and Gordon
  Wetzstein.
\newblock Implicit neural representations with periodic activation functions.
\newblock {\em Advances in Neural Information Processing Systems}, 33, 2020.

\bibitem{song2017semantic}
Shuran Song, Fisher Yu, Andy Zeng, Angel~X Chang, Manolis Savva, and Thomas
  Funkhouser.
\newblock Semantic scene completion from a single depth image.
\newblock In {\em Proc. IEEE Conf. on Computer Vision and Pattern Recognition
  (CVPR)}. IEEE, 2017.

\bibitem{tancik2020fourfeat}
Matthew Tancik, Pratul~P. Srinivasan, Ben Mildenhall, Sara Fridovich-Keil,
  Nithin Raghavan, Utkarsh Singhal, Ravi Ramamoorthi, Jonathan~T. Barron, and
  Ren Ng.
\newblock Fourier features let networks learn high frequency functions in low
  dimensional domains.
\newblock {\em NeurIPS}, 2020.

\bibitem{tchapmi2019topnet}
Lyne~P Tchapmi, Vineet Kosaraju, Hamid Rezatofighi, Ian Reid, and Silvio
  Savarese.
\newblock Topnet: Structural point cloud decoder.
\newblock In {\em Proceedings of the IEEE Conference on Computer Vision and
  Pattern Recognition}, pages 383--392, 2019.

\bibitem{Wang_2020_CVPR}
Xiaogang Wang, Marcelo H. Ang~Jr.  , and Gim~Hee Lee.
\newblock Cascaded refinement network for point cloud completion.
\newblock In {\em Proceedings of the IEEE/CVF Conference on Computer Vision and
  Pattern Recognition (CVPR)}, June 2020.

\bibitem{wang2020self}
Xiaogang Wang, Marcelo~H Ang~Jr, and Gim~Hee Lee.
\newblock A self-supervised cascaded refinement network for point cloud
  completion.
\newblock {\em arXiv preprint arXiv:2010.08719}, 2020.

\bibitem{wang2019forknet}
Yida Wang, David~Joseph Tan, Nassir Navab, and Federico Tombari.
\newblock Forknet: Multi-branch volumetric semantic completion from a single
  depth image.
\newblock In {\em Proceedings of the IEEE International Conference on Computer
  Vision}, pages 8608--8617, 2019.

\bibitem{wang_softpool}
Yida Wang, David~Joseph Tan, Nassir Navab, and Federico Tombari.
\newblock Softpoolnet: Shape descriptor for point cloud completion and
  classification.
\newblock In Andrea Vedaldi, Horst Bischof, Thomas Brox, and Jan-Michael Frahm,
  editors, {\em Computer Vision -- ECCV 2020}, pages 70--85, Cham, 2020.
  Springer International Publishing.

\bibitem{Wen_2020_CVPR}
Xin Wen, Tianyang Li, Zhizhong Han, and Yu-Shen Liu.
\newblock Point cloud completion by skip-attention network with hierarchical
  folding.
\newblock In {\em Proceedings of the IEEE/CVF Conference on Computer Vision and
  Pattern Recognition (CVPR)}, June 2020.

\bibitem{wen2020pmp}
Xin Wen, Peng Xiang, Zhizhong Han, Yan-Pei Cao, Pengfei Wan, Wen Zheng, and
  Yu-Shen Liu.
\newblock Pmp-net: Point cloud completion by learning multi-step point moving
  paths.
\newblock {\em arXiv preprint arXiv:2012.03408}, 2020.

\bibitem{wu2020scfusion}
Shun-Cheng Wu, Keisuke Tateno, Nassir Navab, and Federico Tombari.
\newblock Scfusion: Real-time incremental scene reconstruction with semantic
  completion.
\newblock {\em arXiv preprint arXiv:2010.13662}, 2020.

\bibitem{wu2019pointconv}
Wenxuan Wu, Zhongang Qi, and Li Fuxin.
\newblock Pointconv: Deep convolutional networks on 3d point clouds.
\newblock In {\em Proceedings of the IEEE Conference on Computer Vision and
  Pattern Recognition}, pages 9621--9630, 2019.

\bibitem{xia2021asfm}
Yaqi Xia, Yan Xia, Wei Li, Rui Song, Kailang Cao, and Uwe Stilla.
\newblock Asfm-net: Asymmetrical siamese feature matching network for point
  completion.
\newblock {\em arXiv preprint arXiv:2104.09587}, 2021.

\bibitem{grnet_xie}
Haozhe Xie, Hongxun Yao, Shangchen Zhou, Jiageng Mao, Shengping Zhang, and
  Wenxiu Sun.
\newblock Grnet: Gridding residual network for dense point cloud completion.
\newblock In Andrea Vedaldi, Horst Bischof, Thomas Brox, and Jan-Michael Frahm,
  editors, {\em Computer Vision -- ECCV 2020}, pages 365--381, Cham, 2020.
  Springer International Publishing.

\bibitem{DISN19}
Qiangeng Xu, Weiyue Wang, Duygu Ceylan, Radomir Mech, and Ulrich Neumann.
\newblock Disn: Deep implicit surface network for high-quality single-view 3d
  reconstruction.
\newblock In H. Wallach, H. Larochelle, A. Beygelzimer, F. d\textquotesingle
  Alch\'{e}-Buc, E. Fox, and R. Garnett, editors, {\em Advances in Neural
  Information Processing Systems 32}, pages 492--502. Curran Associates, Inc.,
  2019.

\bibitem{yang2018dense}
Bo Yang, Stefano Rosa, Andrew Markham, Niki Trigoni, and Hongkai Wen.
\newblock Dense 3d object reconstruction from a single depth view.
\newblock {\em IEEE transactions on pattern analysis and machine intelligence},
  2018.

\bibitem{yang20173d}
Bo Yang, Hongkai Wen, Sen Wang, Ronald Clark, Andrew Markham, and Niki Trigoni.
\newblock 3d object reconstruction from a single depth view with adversarial
  learning.
\newblock In {\em Proceedings of the IEEE International Conference on Computer
  Vision Workshops}, pages 679--688, 2017.

\bibitem{yang2018foldingnet}
Yaoqing Yang, Chen Feng, Yiru Shen, and Dong Tian.
\newblock Foldingnet: Point cloud auto-encoder via deep grid deformation.
\newblock In {\em Proceedings of the IEEE Conference on Computer Vision and
  Pattern Recognition}, pages 206--215, 2018.

\bibitem{yu2021pointr}
Xumin Yu, Yongming Rao, Ziyi Wang, Zuyan Liu, Jiwen Lu, and Jie Zhou.
\newblock Pointr: Diverse point cloud completion with geometry-aware
  transformers.
\newblock In {\em Proceedings of the IEEE/CVF International Conference on
  Computer Vision}, pages 12498--12507, 2021.

\bibitem{yuan2018pcn}
Wentao Yuan, Tejas Khot, David Held, Christoph Mertz, and Martial Hebert.
\newblock Pcn: Point completion network.
\newblock In {\em 2018 International Conference on 3D Vision (3DV)}, pages
  728--737. IEEE, 2018.

\bibitem{zhang2020point}
Junming Zhang, Weijia Chen, Yuping Wang, Ram Vasudevan, and Matthew
  Johnson-Roberson.
\newblock Point set voting for partial point cloud analysis.
\newblock {\em arXiv preprint arXiv:2007.04537}, 2020.

\bibitem{zhang2019cascaded}
Pingping Zhang, Wei Liu, Yinjie Lei, Huchuan Lu, and Xiaoyun Yang.
\newblock Cascaded context pyramid for full-resolution 3d semantic scene
  completion.
\newblock In {\em Proceedings of the IEEE/CVF International Conference on
  Computer Vision}, pages 7801--7810, 2019.

\bibitem{zhang2020detail}
Wenxiao Zhang, Qingan Yan, and Chunxia Xiao.
\newblock Detail preserved point cloud completion via separated feature
  aggregation.
\newblock {\em arXiv preprint arXiv:2007.02374}, 2020.

\end{thebibliography}
}

\newpage

\section{Supplementary materials}
As we discussed in the paper, this document aims at showing the detailed parameters of our architectures and more comprehensive results for both object completion and semantic scene completion. It also includes additional qualitative results that compares different methods against the proposed.

\subsection{Parameters in architectures}

This work introduces two architectures to highlight the benefits of the proposed layers.
We list the parameters set in every layer of our direct architecture in \tabref{tab:architecture_direct} and our transformer architecture in \tabref{tab:architecture_transformer}.

\begin{table}[!b]
\centering
\includegraphics[width=0.8\linewidth]{CVPR_direct_architecture.pdf} 
\resizebox{0.9\linewidth}{!}
{
\begin{tabular}{r|ccc}
\toprule	
Layers & \multicolumn{3}{l}{Parameters} \\
\midrule
Feature Extraction & $s = 10$ & $D_\text{out} = 16$ \\
Neighbor Pooling & $\tau = 8$  \\
Feature Extraction & $s = 10$ & $D_\text{out} = 64$ \\
Neighbor Pooling & $\tau = 4$ \\
Feature Extraction & $s = 10$ & $D_\text{out} = 64$ \\
Neighbor Pooling & $\tau = 4$ \\
Feature Extraction & $s = 10$ & $D_\text{out} = 64$ \\
Max Pooling & -- \\
\midrule
Up-Sampling & $s = 5$ & $N_\text{up} = 2$ & $D_\text{out} = 256$ \\
Up-Sampling & $s = 10$ & $N_\text{up} = 8$ & $D_\text{out} = 64$ \\
Up-Sampling & $s = 10$ & $N_\text{up} = 4$ & $D_\text{out} = 64$ \\
Up-Sampling & $s = 10$ & $N_\text{up} = 4$ & $D_\text{out} = 32$ \\
Up-Sampling & $s = 10$ & $N_\text{up} = 8$ & $D_\text{out} = 3$ \\
Up-Sampling & $s = 10$ & $N_\text{up} = 1$ & $D_\text{out} = 3$ \\
Up-Sampling & $s = 10$ & $N_\text{up} = 8$ & $D_\text{out} = 3$ \\
\bottomrule
\end{tabular}
}
\caption{Parameters in each layer of our \emph{direct} architecture.
\label{tab:architecture_direct}
}
\end{table}

\begin{table}[!b]
\centering
\includegraphics[width=0.8\linewidth]{CVPR_transformer_architecture.pdf} 
\resizebox{0.9\linewidth}{!}
{
\begin{tabular}{r|ccc}
\toprule	
Layers & \multicolumn{3}{l}{Parameters} \\
\midrule
Feature Extraction & $s = 10$ & $D_\text{out} = 16$ & \\
Neighbor Pooling & $\tau = 4$ \\
Feature Extraction & $s = 10$ & $D_\text{out} = 64$ & \\
Neighbor Pooling & $\tau = 4$ \\
Positional Coding & -- & & \\
\midrule
Transformer & \multicolumn{2}{l}{\textit{Similar to \cite{yu2021pointr}}} \\
\midrule
Feature Extraction & $s = 10$ & $D_\text{out} = 64$ \\
Up-Sampling & $s = 10$ & $N_\text{up} = 8$  & $D_\text{out} = 3$ \\
Feature Extraction & $s = 10$ & $D_\text{out} = 64$ \\
Feature Extraction & $s = 10$ & $D_\text{out} = 64$ \\
Up-Sampling & $s = 10$ & $N_\text{up} = 8$&$D_\text{out} = 3$ \\
\bottomrule
\end{tabular}
}
\caption{Parameters in each layer of our \emph{transformer} architecture.
\label{tab:architecture_transformer}
}
\end{table}

\subsection{Object completion}

We exhibit a more detailed comparison on the object completion evaluation in \tabref{tab:shapenet_2k_L2}, \tabref{tab:shapenet_16k_L1} and \tabref{tab:shapenet_16k_F1} for the Completion3D~\cite{tchapmi2019topnet}, PCN~\cite{yuan2018pcn} and MVP~\cite{pan2021variational} datasets, respectively.
While we only show the average results in the paper, these tables show the per-category evaluation. 
Based on these results, our architectures are better in most categories when evaluating the Chamfer distance in
\tabref{tab:shapenet_2k_L2} and  \tabref{tab:shapenet_16k_L1}; while, better in all categories when evaluating the F-Score in \tabref{tab:shapenet_16k_F1}.

\subsection{Semantic scene completion with voxels}

Since most of the point cloud approaches only perform completion, we compared our semantic scene completion results to the voxel-based approaches in \tabref{tab:nyu}.
In order to do this, we converted our high resolution point cloud to a lower resolution $60 \times 36 \times 60$ voxels. 
\tabref{tab:nyu} shows the per-category comparison against the voxel-based approaches. Notably, although  downsizing our point cloud introduces errors and difference (\eg the objects in the point cloud are hollow while in the voxels are solid), we still achieve competitive IoU results.

\subsection{Semantic scene completion with point clouds}

\begin{figure*}[!b]
\centering
\includegraphics[width=1.0\linewidth]{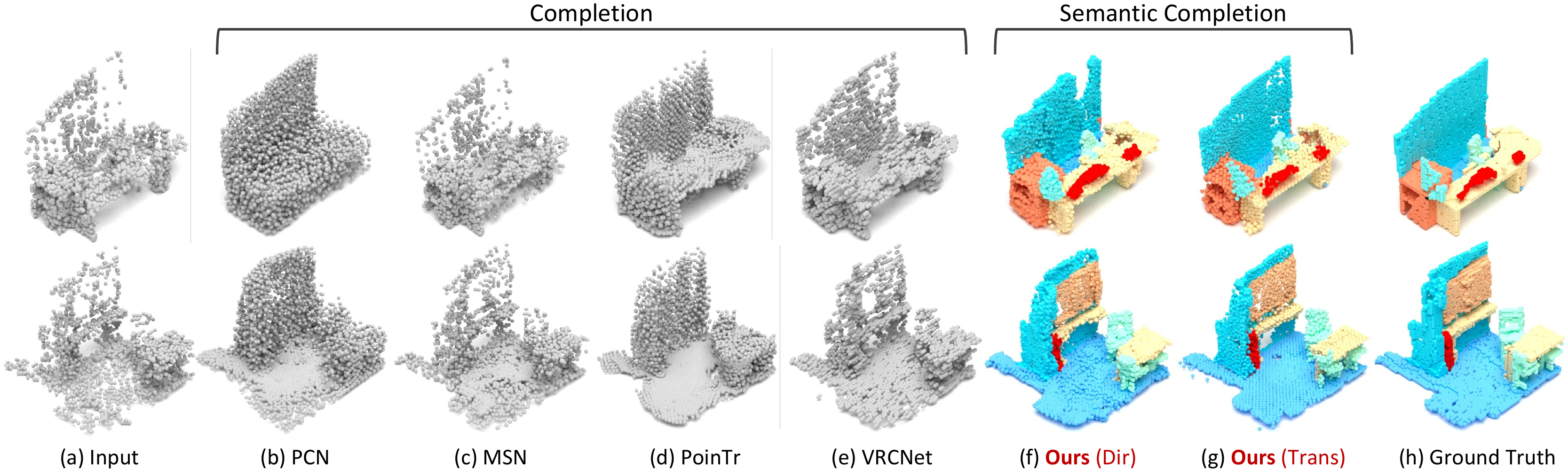}
\caption{Semantic scene completion results on the
CompleteScanNet~\cite{wu2020scfusion} dataset}
\label{fig:scene_qualitatives}
\end{figure*}

We illustrate the semantic scene completion results in \figref{fig:scene_qualitatives}, evaluated on CompleteScanNet~\cite{wu2020scfusion}.
Since there is no other point cloud completion approach that explicitly claim that they can reconstruct scenes, we utilize the architectures that were designed for object completion: 
PCN~\cite{yuan2018pcn}, 
MSN~\cite{liu2020morphing}, 
PoinTr~\cite{yu2021pointr} and
VRCNet~\cite{pan2021variational}.
Due to this, in \figref{fig:scene_qualitatives}, we perform the more complicated semantic completion while the other methods carry out the simpler completion task. 

We observe from the other methods~\cite{yuan2018pcn,liu2020morphing,yu2021pointr,pan2021variational} that their results show a high level of noise such that the objects in the scenes are no longer comprehensible. 
In comparison, our results have significantly less noise and produce reconstructions that are very similar to the ground truth. 
Moreover, a particular attention is given to PoinTr~\cite{yu2021pointr} since we derived our transformer architecture from them. 
Comparing our results against \cite{yu2021pointr}, our reconstructions are significantly more accurate. This in effect demonstrate the important contribution of our proposed layers to our transformer architecture.

\begin{table*}[!ht]
\centering
\begin{tabular}{l|cccccccc|c}
\multicolumn{7}{l}{Output Resolution = 2,048, L2 metric, Completion3D~\cite{tchapmi2019topnet} benchmark} \\
\toprule	
 \multicolumn{1}{c}{Method} 
  & plane & cabinet & car & chair & lamp & sofa & table & vessel & \emph{Avg.} \\
\midrule 
       FoldingNet~\cite{yang2018foldingnet} & 12.83 & 23.01 & 14.88 & 25.69 & 21.79 & 21.31 & 20.71 & 11.51 & 19.07 \\
       PointSetVoting~\cite{zhang2020point} & 6.88 & 21.18 & 15.78 & 22.54 & 18.78 & 28.39 & 19.96 & 11.16 & 18.18 \\
       AtlasNet~\cite{Groueix_2018_CVPR} & 10.36 & 23.40 & 13.40 & 24.16 & 20.24 & 20.82 & 17.52 & 11.62 & 17.77 \\
       PCN~\cite{yuan2018pcn} & 9.79 & 22.70 & 12.43 & 25.14 & 22.72 & 20.26 & 20.27 & 11.73 & 18.22 \\
       TopNet~\cite{tchapmi2019topnet} & 7.32 & 18.77 & 12.88 & 19.82 & 14.60 & 16.29 & 14.89 & 8.82 & 14.25 \\
       SA-Net~\cite{Wen_2020_CVPR} & 5.27 & 14.45 & 7.78 & 13.67 & 13.53 & 14.22 & 11.75 & 8.84 & 11.22 \\
       SoftPoolNet~\cite{wang_softpool} & 6.39 & 17.26 & 8.72 & 13.16 & 10.78 & 14.95 & 11.01 & 6.26 & 11.07 \\
       GRNet~\cite{grnet_xie} & 6.13 & 16.90 & 8.27 & 12.23 & 10.22 & 14.93 & 10.08 & 5.86 & 10.64 \\
       PMP-Net~\cite{wen2020pmp} & 3.99 & 14.70 & 8.55 & 10.21 & 9.27 & 12.43 & 8.51 & 5.77 & 9.23 \\
       CRN~\cite{Wang_2020_CVPR} & 3.38 & 13.17 & 8.31 & 10.62 & 10.00 & 12.86 & 9.16 & 5.80 & 9.21 \\
       SCRN~\cite{wang2020self} & 3.35 & 12.81 & 7.78 & 9.88 & 10.12 & 12.95 & 9.77 & 6.10 & 9.13 \\
       VRCNet~\cite{pan2021variational} & 3.94 & 10.93 & 6.44 & 9.32 & 8.32 & 11.35 & 8.60 & 5.78 & 8.12 \\
       ASFM-Net~\cite{xia2021asfm} & \textbf{2.38} & 9.68 & 5.84 & \textbf{7.47} & 7.11 & \textbf{9.65} & \textbf{6.25} & 4.84 & 6.68 \\
\midrule 
       Ours (direct) & 3.52 & 12.72 & 7.37 & 9.21 & 8.57 & 11.66 & 8.77 & 4.97 & 8.35 \\
       --without $\mathcal{L}_\text{order}$ & 3.64 & 12.83 & 7.48 & 9.34 & 8.70 & 11.79 & 8.88 & 5.07 & 8.47 \\
       Ours (transformer) & 2.41 & \textbf{9.54} & \textbf{4.99} & 7.89 & \textbf{6.89} & 9.92 & 7.20 & \textbf{4.29} & \textbf{6.64} \\
       --without $\mathcal{L}_\text{order}$ & 2.48 & 9.62 & 5.10 & 7.99 & 7.01 & 10.04 & 7.29 & 4.39 & 6.74 \\
\bottomrule
\end{tabular}
\caption{Evaluation on the object completion on Completion3D~\cite{tchapmi2019topnet} benchmark based on the Chamfer distance trained with L2 distance (multiplied by $10^4$) with the output resolution of 2,048.
}
 \label{tab:shapenet_2k_L2}
\end{table*}

\begin{table*}[!t]
\centering
\begin{tabular}{l|cccccccc|c}
\multicolumn{6}{l}{Output Resolution = 16,384, L1 metric, PCN~\cite{yuan2018pcn} dataset} \\
\toprule	
 \multicolumn{1}{c}{Method} 
 & plane & cabinet & car & chair & lamp & sofa & table & vessel & \emph{Avg.} \\
\midrule 
       3D-EPN~\cite{dai2017shape} & 13.16 & 21.80 & 20.31 & 18.81 & 25.75 & 21.09 & 21.72 & 18.54 & 20.15 \\
       ForkNet~\cite{wang2019forknet} & 9.08 & 14.22 & 11.65 & 12.18 & 17.24 & 14.22 & 11.51 & 12.66 & 12.85 \\
\midrule 
       PointNet++~\cite{qi2017pointnet++} & 10.30 & 14.74 & 12.19 & 15.78 & 17.62 & 16.18 & 11.68 & 13.52 & 14.00 \\
       FoldingNet~\cite{yang2018foldingnet} & 9.49 & 15.80 & 12.61 & 15.55 & 16.41 & 15.97 & 13.65 & 14.99 & 14.31 \\
       AtlasNet~\cite{Groueix_2018_CVPR} & 6.37 & 11.94 & 10.11 & 12.06 & 12.37 & 12.99 & 10.33 & 10.61 & 10.85 \\
       TopNet~\cite{tchapmi2019topnet} & 7.61 & 13.31 & 10.90 & 13.82 & 14.44 & 14.78 & 11.22 & 11.12 & 12.15 \\ 
       PCN~\cite{yuan2018pcn} & 5.50 & 10.63 & 8.70 & 11.00 & 11.34 & 11.68 & 8.59 & 9.67 & 9.64 \\
       MSN~\cite{liu2020morphing} & 5.60 & 11.96 & 10.78 & 10.62 & 10.71 & 11.90 & 8.70 & 9.49 & 9.97 \\
       SoftPoolNet~\cite{wang_softpool} & 6.93 & 10.91 & 9.78 & 9.56 & 8.59 & 11.22 & 8.51 & 8.14 & 9.20 \\
       GRNet~\cite{grnet_xie} & 6.45 & 10.37 & 9.45 & 9.41 & 7.96 & 10.51 & 8.44 & 8.04 & 8.83 \\
       PMP-Net~\cite{wen2020pmp} & 5.65 & 11.24 & 9.64 & 9.51 & \textbf{6.95} & 10.83 & 8.72 & 7.25 & 8.73 \\
       CRN~\cite{Wang_2020_CVPR} & 4.79 & 9.97 & 8.31 & 9.49 & 8.94 & 10.69 & 7.81 & 8.05 & 8.51 \\
       SCRN~\cite{wang2020self} & 4.80 & 9.94 & 9.31 & 8.78 & 8.66 & 9.74 & 7.20 & 7.91 & 8.29 \\
       PoinTr~\cite{yu2021pointr} & 4.75 & 10.47 & 8.68 & 9.39 & 7.75 & 10.93 & 7.78 & 7.29 & 8.38 \\
\midrule 
       Ours (direct) & 5.34 & \textbf{9.20} & \textbf{8.26} & 8.96 & 9.40 & \textbf{10.46} & 7.54 & 8.56 & 8.47 \\
       --\textit{without} $\mathcal{L}_\text{order}$ & 5.47 & 9.34 & 8.37 & 9.09 & 9.54 & 10.59 & 7.69 & 8.66 & 8.59 \\
       Ours (transformer) & \textbf{4.43} & 10.03 & 8.28 & \textbf{8.96} & 7.29 & 10.55 & \textbf{7.31} & \textbf{6.85} & \textbf{7.96} \\
       --\textit{without} $\mathcal{L}_\text{order}$ & 4.56 & 10.17 & 8.42 & 9.10 & 7.41 & 10.66 & 7.41 & 6.96 & 8.09 \\
\bottomrule
\end{tabular}
\caption{Evaluation on the object completion on PCN~\cite{yuan2018pcn} dataset based on the Chamfer distance trained with L1 distance (multiplied by $10^3$) with the output resolution of 16,384.
 \label{tab:shapenet_16k_L1}
}
\end{table*}

\begin{table*}[!t]
\centering
\begin{tabular}{l|cccccccc|c}
\multicolumn{7}{l}{Output Resolution = 16,384, F-Score@1\%, MVP~\cite{pan2021variational} dataset} \\
\toprule	
 \multicolumn{1}{c}{Method} 
 & plane & cabinet & car & chair & lamp & sofa & table & vessel & \emph{Avg.} \\
\midrule 
       TopNet~\cite{tchapmi2019topnet} & 0.789 & 0.621 & 0.612 & 0.443 & 0.387 & 0.506 & 0.639 & 0.609 & 0.576 \\ 
       PCN~\cite{yuan2018pcn} & 0.816 & 0.614 & 0.686 & 0.517 & 0.455 & 0.552 & 0.646 & 0.628 & 0.614 \\
       MSN~\cite{liu2020morphing} & 0.879 & 0.692 & 0.693 & 0.599 & 0.604 & 0.627 & 0.730 & 0.696 & 0.690 \\
       SoftPoolNet~\cite{wang_softpool} & 0.843 & 0.568 & 0.636 & 0.623 & 0.698 & 0.568 & 0.680 & 0.71 & 0.666 \\
       GRNet~\cite{grnet_xie} & 0.853 & 0.578 & 0.646 & 0.635 & 0.710 & 0.580 & 0.690 & 0.723 & 0.677 \\
       ECG~\cite{pan2020ecg} & 0.906 & 0.680 & 0.716 & 0.683 & 0.734 & 0.651 & 0.766 & 0.753 & 0.736 \\
       NSFA~\cite{zhang2020detail} & 0.903 & 0.694 & 0.721 & 0.737 & 0.783 & 0.705 & 0.817 & 0.799 & 0.770 \\
       CRN~\cite{Wang_2020_CVPR} & 0.898 & 0.688 & 0.725 & 0.670 & 0.681 & 0.641 & 0.748 & 0.742 & 0.724 \\
       VRCNet~\cite{pan2021variational} & 0.928 & 0.721 & 0.756 & 0.743 & 0.789 & 0.696 & 0.813 & 0.800 & 0.781 \\
       PoinTr~\cite{yu2021pointr} & 0.888 & 0.681 & 0.716 & 0.703 & 0.749 & 0.656 & 0.773 & 0.760 & 0.741 \\
\midrule 
       Ours (direct) & 0.926 & 0.738 & 0.766 & 0.783 & 0.837 & 0.709 & 0.829 & 0.821 & 0.801 \\
       --without $\mathcal{L}_\text{order}$ & 0.910 & 0.750 & 0.741 & 0.734 & 0.835 & 0.715 & 0.839 & 0.783 & 0.788 \\
       Ours (transformer) & \textbf{0.942} & \textbf{0.753} & \textbf{0.780} & \textbf{0.799} & \textbf{0.851} & \textbf{0.725} & \textbf{0.844} & \textbf{0.836} & \textbf{0.816} \\
       --without $\mathcal{L}_\text{order}$ & 0.922 & 0.731 & 0.759 & 0.776 & 0.831 & 0.703 & 0.824 & 0.813 & 0.795 \\
\bottomrule
\end{tabular}
\caption{Evaluation on the object completion on MVP~\cite{pan2021variational} dataset based on the F-Score@1\% trained with L2 Chamfer distance and the output resolution of 16,384.
 \label{tab:shapenet_16k_F1}
}
\end{table*}


\begin{table*}[!t]
\centering
\resizebox{\textwidth}{!}
{
\begin{tabular}{l|c|c|ccccccccccc|c}
\toprule	
 \multicolumn{1}{c}{Method} 
 & res. & whole & ceil. & floor & wall & win. & chair & bed & sofa & table & tvs & furn. & objs & \emph{Avg.} \\
\midrule 
    Lin \etal~\cite{lin2013holistic} & 60 & 36.4 & 0.0 & 11.7 & 13.3 & 14.1 & 9.4 & 29.0 & 24.0 & 6.0 & 7.0 & 16.2 & 1.1 & 12.0 \\
    Geiger and Wang~\cite{geiger2015joint} & 60 & 44.4 & 10.2 & 62.5 & 19.1 & 5.8 & 8.5 & 40.6 & 27.7 & 7.0 & 6.0 & 22.6 & 5.9 & 19.6 \\
    SSCNet~\cite{song2017semantic} & 60 & 55.1 & 15.1 & 94.6 & 24.7 & 10.8 & 17.3 & 53.2 & 45.9 & 15.9 & 13.9 & 31.1 & 12.6 & 30.5 \\
    VVNet~\cite{guo2018view} & 60 & 61.1 & 19.3 & 94.8 & 28.0 & 12.2 & 19.6 & 57.0 & 50.5 & 17.6 & 11.9 & 35.6 & 15.3 & 32.9 \\
    SaTNet~\cite{liu2018nips} & 60 & 60.6 & 17.3 & 92.1 & 28.0 & 16.6 & 19.3 & 57.5 & 53.8 & 17.7 & 18.5 & 38.4 & 18.9 & 34.4 \\
    ForkNet~\cite{wang2019forknet} & 80 & 37.1 & 36.2 & 93.8 & 29.2 & 18.9 & 17.7 & 61.6 & 52.9 & 23.3 & 19.5 & 45.4 & 20.0 & 37.1 \\
    CCPNet~\cite{zhang2019cascaded} & 240 & 63.5 & 23.5 & 96.3 & 35.7 & 20.2 & 25.8 & 61.4 & 56.1 & 18.1 & 28.1 & 37.8 & 20.1 & 38.5 \\
    SketchSSC~\cite{chen20203d} & 60 & 71.3 & 43.1 & 93.6 & 40.5 & 24.3 & 30.0 & 57.1 & 49.3 & 29.2 & 14.3 & 42.5 & 28.6 & 41.1 \\
    SISNet~\cite{cai2021semantic} & 60 & \textbf{78.2} & \textbf{54.7} & 93.8 & \textbf{53.2} & \textbf{41.9} & \textbf{43.6} & \textbf{66.2} & \textbf{61.4} & \textbf{38.1} & \textbf{29.8} & \textbf{53.9} & \textbf{40.3} & \textbf{52.4} \\
\midrule 
    Ours (direct) & 60 & 63.7 & 38.1 & 97.1 & 37.0 & 15.5 & 18.7 & 55.2 & 54.9 & 29.6 & 21.4 & 49.2 & 23.7 & 40.0 \\ 
    \small{--\textit{with} $\gamma=1$ \textit{in} $\mathcal{L}_\text{semantic}$} & 60 & 58.2 & 35.1 & 94.3 & 34.0 & 12.7 & 15.8 & 52.3 & 52.0 & 26.7 & 18.4 & 46.3 & 20.9 & 37.2 \\
    Ours (transformer) & 60 & 66.1 & 40.4 & \textbf{98.6} & 39.6 & 18.1 & 21.2 & 57.5 & 57.0 & 31.9 & 23.5 & 51.3 & 26.4 & 42.4 \\ 
    \small{--\textit{with} $\gamma=1$ \textit{in} $\mathcal{L}_\text{semantic}$} & 60 & 63.4 & 36.6 & 95.0 & 36.6 & 14.8 & 18.1 & 53.9 & 53.4 & 28.8 & 20.1 & 47.8 & 22.5 & 38.9 \\
\bottomrule
\end{tabular}
}
\caption{Semantic completion on NYU dataset. The value in res.~($x$) is the output volumetric resolution which is $x \times 0.6x \times x$.
 \label{tab:nyu}
}
\end{table*}

\end{document}